\documentclass[preprint,12pt,numbers,sort&compress]{elsarticle}

\usepackage{amsmath} 
\usepackage{amssymb}  
\usepackage{threeparttable}
\usepackage{xcolor}
\usepackage{fancyhdr}
\usepackage{float}
\usepackage{colortbl}
\usepackage{dcolumn}
\usepackage{siunitx}
\usepackage{diagbox}
\usepackage{graphicx}
\usepackage{caption}
\usepackage{amsthm} 
 
\theoremstyle{definition} 
\newtheorem{myTheo}{Theorem}
\newtheorem{myDef}{Definition}

\newtheorem{myRem}{Remark}
\newtheorem{myAss}{Assumption}

\usepackage{threeparttable}
\usepackage[dvipsnames]{xcolor}

\usepackage[linesnumbered,ruled]{algorithm2e}
\renewcommand{\KwData}{\textbf{Hyperparameters:\:}}
\usepackage{rotating} 

\usepackage[colorlinks=true,linkcolor=blue,citecolor=blue,urlcolor=blue]{hyperref}

\journal{Nuclear Physics B}

\begin{document}

\begin{frontmatter}


\title{Deep Deterministic Policy Gradient with Symmetric Data Augmentation for Lateral Attitude Tracking Control of a Fixed-wing Aircraft}

\author[aff1]{Yifei Li\corref{cor1}}
\ead{y.li-34@tudelft.nl}

\author[aff1]{Erik-Jan van Kampen}

\cortext[cor1]{Corresponding author}

\affiliation[aff1]{organization={Section Control and Simulation, Delft University of Technology},
             addressline={Kluyverweg 1},
             city={Delft},
             postcode={2629 HS},
             state={South Holland},
             country={The Netherlands}}

\begin{abstract}
The symmetry of dynamical systems can be exploited for state-transition prediction and to facilitate control policy optimization. This paper leverages system symmetry to develop sample-efficient offline reinforcement learning (RL) approaches. Under the symmetry assumption for a Markov Decision Process (MDP), a symmetric data augmentation method is proposed. The augmented samples are integrated into the dataset of Deep Deterministic Policy Gradient (DDPG) to enhance its coverage of the state space. Furthermore, sample utilization efficiency is improved by introducing a second critic trained on the augmented samples, resulting in a dual-critic structure. The aircraft’s model is verified to be symmetric, and flight control simulations demonstrate accelerated policy convergence when augmented samples are employed.
\end{abstract}


\end{frontmatter}



\section{Introduction}
A common characteristic in the motion of mechanical systems, such as aircraft \cite{Zipfel197613}, cars \cite{Yao2013ITSC00} and robotic arms \cite{Amadio20194}, is symmetry, which arises from the symmetric design of their mechanical structures. This characteristic implies that each state trajectory has a symmetric counterpart with respect to a reference plane, and are associated with corresponding symmetric control polices. As a result, the motion of one state trajectory can be leveraged to infer the motion of its symmetric counterpart. An illustrative example is given as the symmetric motions in a cart-pole system \cite{Mahajan201700}. 

The configurational symmetries of an aircraft frame exhibit two primary types: reflectional and tetragonal symmetries \cite{Zipfel197613}. These structural properties have been exploited to derive conditions under which certain aerodynamic derivatives vanish. They also give rise to kinematic and dynamic symmetries. As a result, aircraft motions exhibit symmetry, which can potentially be leveraged to enhance the learning performance of the RL-based flight control frameworks. In recent years, RL has been successfully applied to flight control design of various aerial vehicles, including quadrotors \cite{Han20227,Huang20235,Wang202050,Ma202372}, fixed-wing aircraft \cite{Chowdhury202460,Bohn202435,Jiang202359} and vertical take-off and landing (VTOL) aircraft \cite{BMa202324}, reducing the dependence on accurate models. In these examples, the flight control law is typically parameterized by a neural network and trained using samples. The quality of a flight dataset, particularly its coverage of the state-action space, directly influences the effectiveness of the learned control law. During the offline training phase, the flight data is generated through the agent interacting with a simulation model, using an exploration policy. However, the convergence of control policy decreases exploration, which may lead to insufficient coverage of the state-action space. This highlights a well-known \textit{exploration and exploitation} trade-off during the learning process \cite{Sutton201800}. Moreover, inadequate design of the exploration policy can exacerbate this issue by limiting the diversity of collected samples, potentially degrading control performance due to poor generalization of the actor over unvisited regions of the state space. These challenges bring up to the issue of \textit{sample efficiency} during learning. To address this issue, data augmentation techniques have been explored to generate additional training samples based on system properties, such as symmetry \cite{Mishra201900,Angelotti202214,Angelotti202315,Brandstetter202239,Wang2021ICLR00,Wang202239,Lin2019NeurIPS00,Lin20205,Pinneri202300}, leading to \textit{symmetric data augmentation}. A motivation of using symmetric data augmentation for RL-based flight control design is the costly exploration of the high-dimensional state-action space of the aircraft dynamics.

This paper begins by formulating the $Q$ function-based policy optimization method. The symmetry of samples is then defined within an MDP framework. A symmetry condition is derived to determine whether samples are symmetric. This theoretical result motivates a symmetric data augmentation method. Furthermore, the augmented samples are incorporated into the Deep Deterministic Policy Gradient (DDPG) algorithm \cite{Lillicrap2016ICLR00}, resulting in a richer dataset. To enhance sample utilization efficiency, we propose a two-step approximate value iteration method. In the first step, the explored samples are used to train a critic and an actor. In the second step, the augmented (symmetric) samples are used to train another critic while updating the same actor. Finally, the symmetry analysis of the aircraft model validates the application of symmetric data augmentation for offline flight control learning.

The main contributions of this paper are summarized as:
\begin{itemize}
    \item A symmetric data augmentation method is proposed to generate additional training samples.
    \item A two-step approximate policy iteration method is developed to enhance sample utilization efficiency during training.
    \item The symmetry of the aircraft’s model is analyzed, and symmetry-informed RL algorithms are applied for flight control to enable sample-efficient policy learning.
\end{itemize}

This paper is structured as follows. Section \ref{section_foundations} formulates the discrete-time optimal control problem and the approximate value iteration method. Section \ref{Section_symmetric_dynamic_model} defines the symmetry characteristics of a dynamical system. Section \ref{section_ddpg_sda} introduces the symmetric data augmentation method. Section \ref{section_ddpg_sca} develops the DDPG with symmetric critic augmentation. Section \ref{section_model} presents the aircraft dynamical model and analyzes its symmetry. Section \ref{section_action_smoothness} introduces action smoothness techniques. Section \ref{section_simulation} provides simulation results of symmetry-informed RL for flight control. Section \ref{section_conclusion} concludes this paper.

\subsection{Related Work}
\subsubsection{Symmetric data augmentation for RL} Symmetric data augmentation relies on the assumption of symmetry throughout the state–action space. This assumption implies that each state trajectory has a symmetric counterpart, allowing samples from an explored trajectory to be used to generate samples from its symmetric counterpart, without further interaction with the environment. This method increases the number of training samples, thereby improving sample efficiency, which is an especially important advantage when applying RL to complex, high-dimensional systems where exploration of the state–action space is costly. In \cite{Mishra201900}, the explored samples are mirrored according to a simple symmetric relation and incorporated into maximum a posterior policy optimization to accelerate learning. In \cite{Angelotti202214,Angelotti202315}, an expert-guided detection method is developed to verify symmetry by assuming system dynamics are invariant. The detected symmetry is used to augment samples to learn the transition functions of Grid, Cart-pole and Acrobot systems. In \cite{Brandstetter202239}, the Lie point symmetry group is used to augment samples for solving a neural partial differential equation (PDE). In \cite{Wang2021ICLR00,Wang202239}, the generalization bounds for data augmentation method is theoretically analyzed using symmetry. In \cite{Lin2019NeurIPS00,Lin20205}, symmetric data augmentation is combined with experience replay techniques, leading to Kaleidoscope Experience Replay based on reflectional symmetry, and Goal-augmented Experience Replay, a direct generalization of Hindsight Experience Replay. Reference \cite{Pinneri202300} learns the equivariant set of an MDP and uses it for data augmentation in offline RL. Reference \cite{Weissenbacher202239} employs a pre-trained forward model for Koopman latent representation of equivariant dynamical systems, enabling symmetric data augmentation for $Q$-learning.

\subsubsection{DDPG/TD3 for flight control} DDPG and twin-delayed DDPG (TD3) are deterministic, off-policy reinforcement learning approaches, enabling policy optimization through a state-action ($Q$) value function over the continuous state–action space of aircraft dynamics. In flight control designs, these methods reduce reliance on accurate system models by leveraging samples and thus mitigate the challenges associated with modeling aerodynamic coefficients, actuator dynamics, and other unknown dynamics. Comprehensive surveys of related research can be found in \cite{Razzaghi2024136,Richter202413}. Specific applications include attitude control design for the Flying-V \cite{Volker2023AIAA00}, Cessna Citation PH-LAB \cite{Gavra202447}, quadrotors \cite{Hu20248,Liu202212,Liu20226,Mysore2021ICRA00,Li202229, Han20227,Liu202252,Chen2022AGNC00}, the F-16 model \cite{DeMarco2023111}, jetstream \cite{Budiarti2019}, fixed-wing unmanned aerial vehicles (UAV) \cite{Tang202437,Xu202237,Li2024ECC00}, UAV automatic carrier landing system \cite{Guo2025TAES00}, the Internet-of-Drones \cite{Li202229}, Skyhunter \cite{Shukla202433}, monocopter \cite{Sufiyan2019ICRA00}, flying-wing aircraft \cite{Song202522}, bionic bird wing-foldable UAV model \cite{Xu201992}, and morphing trailing-edge wing \cite{Duan2024153}. The contribution of this paper lies in the symmetry analysis of the aircraft lateral model and its application to formulating a symmetric data augmentation method. The idea of exploiting aircraft model symmetry to enhance offline policy training has rarely been discussed in the existing literature.

\section{FOUNDATIONS}\label{section_foundations}

This section first formulate an infinite-horizon optimal control problem for a nonlinear system. Then, the state-value ($V$) function and the state–action value ($Q$) function are defined to evaluate the values of states and actions, respectively. Finally, exact value iteration and approximate value iteration are introduced as numerical approaches for solving the optimal control problem.
 
\subsection{Definitions}
Consider a discrete-time control-affine nonlinear system 

\begin{equation}\label{nonlinear_system}
\begin{aligned}
         x_{k+1} &= F(x_{k})x_{k} + G(x_{k}) u_{k}, k\in \mathbb{N}
\end{aligned}
\end{equation}\\
where $x_{k} \in \mathbb{R}^{n}$ and $u_{k} \in \mathbb{R}^{m}$ denote the state and input vectors, respectively. The nonlinear functions $F(x_{k}) \in \mathbb{R}^{n \times n}$ and $G(x_{k}) \in \mathbb{R}^{n \times m}$ are associated with $x_{k}$. The subscript $k$ denotes the time-step index, and $\mathbb{N}$ denotes the set of non-negative integers.

Define a performance index starting from the initial state $x_{0}$ as

\begin{equation}\label{performanceindex}
    J(x_{0},u_{0}) = \sum^{\infty}_{k=0}\gamma^{k} r(x_{k},u_{k})
\end{equation}\\
where $r(x_{k}, u_{k})$ denotes the reward function at step $k$, and $\gamma \in [0,1]$ is the discount factor. The control input sequence $\{u_{k}|k=0,1,2,\cdots\}$ is generated by a state-feedback control law $u_{k} = h(x_{k})$.


\begin{myAss}\label{Assumption_Lipschitz}
   The control policy $h(x)$ is Lipschitz continuous with respect to the state $x$.
\end{myAss}

Assumption \ref{Assumption_Lipschitz} is commonly satisfied in the control design of real-world mechanical systems. This condition ensures the existence of the policy gradient and facilitates gradient-based policy search. A neural network policy can be constructed to satisfy Lipschitz continuity by employing Lipschitz continuous activation functions, such as $\tanh$ and $\text{ReLU}$ \cite{Song2018NeurIPS00,Avant202435,Scaman2018}.



\begin{myDef}
    The state-value function $V^{h}: \mathbb{R}^{n}\rightarrow\mathbb{R}_{+}$, which starts from any initial state $x_{0}$ by following a deterministic policy $h(\cdot)$, is defined by

\begin{equation}\label{Vfun}
    V^{h}(x_{0}) = \sum^{\infty}_{k=0}\gamma^{k}r(x^{h}_{k},h(x^{h}_{k}))
\end{equation}\\
where $x^{h}_{k}$ denotes the state $x_{k}$ following a control policy $h(\cdot)$.
\end{myDef}

\begin{myDef}
    
The state-action value function $Q^{h}: \mathbb{R}^{n+m}\rightarrow \mathbb{R}_{+}$, which starts from any initial state $x_{0}$ by taking an action $a_{0}$ and following $h(\cdot)$, is defined by

\begin{equation}\label{Qfun}
    Q^{h}(x_{0},a_{0}) = r(x_{0},a_{0})+\sum^{\infty}_{t=1}\gamma^{k}r(x^{h}_{k},h(x^{h}_{k}))
\end{equation}
\end{myDef}

The Bellman equations for $V^{h}(x_{0})$ is written as

\begin{equation}\label{bellmanV}
    V^{h}(x_{0}) = r(x_{0},h(x_{0})) + \gamma V^{h}(x_{1})
\end{equation}

The Bellman equations for $Q^{h}(x_{0},a_{0})$ is written as

\begin{equation}\label{bellmaneq}
    Q^{h}(x_{0},a_{0}) = r(x_{0},a_{0}) + \gamma V^{h}(x_{1})
\end{equation}

Use $V^{h}(x_{1})=Q^{h}(x_{1},h(x_{1}))$ in Equation \ref{bellmaneq}:

\begin{equation}\label{Qh}
    Q^{h}(x_{0},a_{0}) = r(x_{0},a_{0}) + \gamma Q^{h}(x_{1},h(x_{1}))
\end{equation}


\subsection{Policy optimization based on Q function}



Define an optimal control policy starting from $x_{0}$ with the policy $h^{*}(x_{0}) = \arg\max_{h(\cdot)}V^{h}(x_{0})$. 

Use $h^{*}(x_{0})$ in Equation \ref{bellmanV}:

\begin{equation}\label{optbellmanV}
    V^{h^{*}}(x_{0}) = \underset{h(\cdot)}{\text{max}}\Big(r(x_{0},h(x_{0})) + \gamma V^{h}(x_{1})\Big)
\end{equation}



The optimal $Q$ function at the state-action pair $(x_{0},a_{0})$ is defined to be following an optimal control policy $h^{*}(\cdot)$ from $x_{1}$:

\begin{equation}
\begin{aligned}
    \Big(Q^{h}(x_{0},a_{0})\Big)^{*} :&=  Q^{h^{*}}(x_{0},a_{0})\\
    &= \max_{h(\cdot)}\Big(r(x_{0},a_{0})+\gamma 
    V^{h}(x_{1})\Big)\\
    &=r(x_{0},a_{0})+\gamma V^{h^{*}}(x_{1})
\end{aligned}    
\end{equation}

The optimal action $a^{*}_{0}$ maximizes $Q^{h}(x_{0},a_{0})$ as

\begin{equation}\label{optact}
\begin{aligned}
        a^{*}_{0}&= \arg\max_{a_{0}}Q^{h}(x_{0},a_{0})\\
    &= \arg\max_{a_{0}}\Big(r(x_{0},a_{0}) +\gamma Q^{h}(x_{1},h(x_{1}))\Big)
\end{aligned}
\end{equation}

Use $a^{*}_{0}$ in Equation \ref{Qh}:
\begin{equation}\label{optQ}
    Q^{h}(x_{0},a^{*}_{0}) = \max_{a_{0}} \Big(r(x_{0},a_{0}) +\gamma Q^{h}(x_{1},h(x_{1}))\Big)
\end{equation}

\subsection{Exact value iteration}
The exact value iteration \cite{Sutton201800} recursively evaluates and improves the current policy that produces actions $a_{k}=h(x_{k}),\forall x_{k}\in\mathcal{S}$, where $\mathcal{S}\in\mathbb{R}^{n}$ denotes the  state space.\\
\textbf{Policy Evaluation}
\begin{equation}\label{bellmaequation}
    (Q^{h})^{i+1}(x_{k},a_{k}) = r(x_{k},a_{k}) + \gamma (Q^{h})^{i}(x_{k+1},h^{i}(x_{k+1})),\; \forall x_{k},x_{k+1}\in\mathcal{S}
\end{equation}\\
\textbf{Policy Improvement}\\
\begin{equation}\label{policyimprovementstep2}
\begin{aligned}
        h^{i+1}(x_{k}) &= \arg\max_{h(\cdot)}(Q^{h})^{i+1}(x_{k},h(x_{k}))
\end{aligned}
\end{equation}\\
where $i\geq0$ denotes the iteration index. The convergence to a fixed point, $\lim_{i \rightarrow \infty} (Q^{h})^{i} = (Q^{h})^{*}$, $\lim_{i \rightarrow \infty} h^{i} = h^{*}$ is proven based on the $\gamma$-contraction property of the Bellman operator \cite{Sutton201800}.

\subsection{Approximate value iteration} \label{Subsection_API}


The functions $Q^{h}(x,a), h(x)$ are usually unknown, so that solving Equations \ref{bellmaequation}, \ref{policyimprovementstep2} are difficult. A common approach is to parameterize these functions using neural networks, leading to an approximate method for solving these equations, namely Approximate Value Iteration (AVI). \cite{Heydari2016}. AVI uses two neural networks: (1) the critic network approximates the state-action value function and is denoted as $\hat{Q}^{h}_{\psi}(x,a)$, with a parameter set $\psi$; (2) The actor network approximates the policy $h(x)$ and is denoted as $\mu_{\vartheta}(x)$, with a parameter set $\vartheta$. Therefore, Equations \ref{bellmaequation}, \ref{policyimprovementstep2} can be reconstructed as\newline

\textbf{Approximate Policy Evaluation}
\begin{equation}\label{approximatedbellmaequation}
    (\hat{Q}^{h}_{\psi})^{i+1}(x_{k},a_{k}) = r(x_{k},a_{k}) + \gamma (\hat{Q}^{h}_{\psi})^{i}(x_{k+1},(\mu_{\vartheta})^{i}(x_{k+1})),\; \forall x_{k},x_{k+1}\in\mathcal{S}
\end{equation}

\textbf{Approximate Policy Improvement}
\begin{equation}\label{policyimprovementstep3}
        (\mu_{\vartheta})^{i+1} (x_{k})= \arg\max_{\vartheta}(\hat{Q}^{h}_{\psi})^{i+1}(x_{k},\mu_{\vartheta}(x_{k}))
\end{equation}

The parameter sets $\psi$ and $\vartheta$ are optimized using gradient descent in the DDPG algorithm \cite{Lillicrap2016ICLR00}.

\section{Symmetric Dynamical Model}\label{Section_symmetric_dynamic_model}
The Markov property of system \ref{nonlinear_system} is assumed to hold and is defined as: the transition of states depends merely on the present states and actions, and not on any previous states \cite{Sutton201800}. Accordingly, a sample of state transition is represented as $(x_{k},a_{k},x_{k+1})$, indicating that the system transitions from $x_{k}$ to $x_{k+1}$ when action $a_{k}$ is applied.

\begin{myDef}
    (Symmetric one-step state transitions) Two state transition samples, denoted as $(x_{k},a_{k},x_{k+1})$ and $(x^{\prime}_{k},a^{\prime}_{k},x^{\prime}_{k+1})$, are symmetric to a reference state $x^{*}$ when 

\begin{equation}\label{symxt}     
        \frac{x_{k}+x^{\prime}_{k}}{2}=x^{*}
\end{equation}

\begin{equation}\label{symat}
    a_{k}=-a^{\prime}_{k}
\end{equation}

\begin{equation}\label{symxt+1}
    \frac{x_{k+1}+x^{\prime}_{k+1}}{2}=x^{*}
\end{equation}
\end{myDef}

Assuming Equations \ref{symxt}, \ref{symat} hold, the following theorem discusses the condition under which Equation \ref{symxt+1} holds.

\begin{myTheo}\label{Theorem_symmetry}
(Symmetry of $x_{k+1}$) Select two samples from the system \ref{nonlinear_system}, denoted as $(x_{k},a_{k},x_{k+1})$, $(x^{\prime}_{k},a^{\prime}_{k},x^{\prime}_{k+1})$, and a reference state $x=x^{*}$. Assuming Equations \ref{symxt}, \ref{symat} hold, then $x_{k+1},x^{\prime}_{k+1}$ are symmetric with respect to $x^{*}$ when\\
(1) $x^{*}=0\in\mathbb{R}^{n},G(x_{k})=G(x^{\prime}_{k}),F(x_{k})=F(x^{\prime}_{k})$\\
(2) $x^{*}\neq0\in\mathbb{R}^{n}, G(x_{k})=G(x^{\prime}_{k}), F(x_{k})=F(x^{\prime}_{k})=I\in\mathbb{R}^{n\times n}$.
\end{myTheo}
\begin{proof}
    See \ref{Appendix_proof_of_theorem_1}. 
\end{proof}

\begin{figure}
\centering
\includegraphics[width=0.62\linewidth]{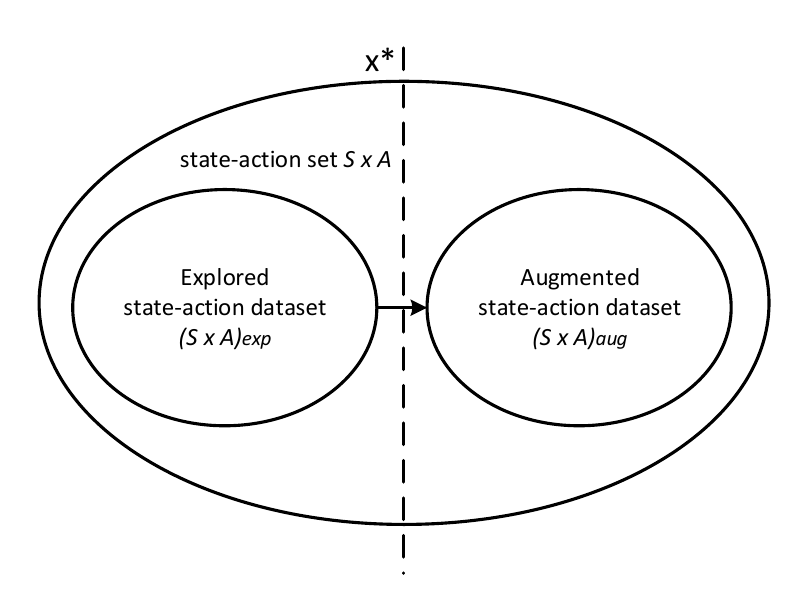}
\caption{A schematic illustration of symmetry in the state–action space $\mathcal{S}\times\mathcal{A}$ of a dynamical system. The full state–action space $\mathcal{S}\times\mathcal{A}$ is composed of two subspaces: (1) the subspace containing explored samples, denoted $(\mathcal{S}\times\mathcal{A})_{\text{exp}}$, and (2) the subspace containing augmented samples, denoted $(\mathcal{S}\times\mathcal{A})_{\text{aug}}$.}
\label{Figure_symmetric_state_action_space}
\end{figure}

\begin{myDef}
    The system \ref{nonlinear_system} is a symmetric dynamical system with respect to a reference state $x^{*}$ when
\begin{equation}
\begin{aligned}
        \frac{x_{k}+x^{\prime}_{k}}{2}=x^{*},\quad &\forall x_{k}, x_{k+1}\in \mathcal{S}\\
    a_{k}=-a^{\prime}_{k},\quad &\forall a_{k},a^{\prime}_{k}\in \mathcal{A}\\
    \frac{x_{k+1}+x^{\prime}_{k+1}}{2}=x^{*},\quad &\forall x^{\prime}_{k},x^{\prime}_{k+1}\in \mathcal{S}
\end{aligned}
\end{equation}\\
where $\mathcal{A}\in\mathbb{R}^{m}$ denotes the action space. This symmetry relation is illustrated in Figure \ref{Figure_symmetric_state_action_space}.
\end{myDef}

\begin{myDef}
     The reward function $r(x_{k},u_{k})$ is symmetric at  state-action pairs $(x_{k},a_{k})$ and $(x^{\prime}_{k},a^{\prime}_{k})$ when 
\begin{equation}\label{symr}
r(x_{k},a_{k})=r(x^{\prime}_{k},a^{\prime}_{k})
\end{equation}
\end{myDef}

\begin{myDef}
The action-value function $Q^{h}(x_{k},a_{k})$ in Equation \ref{Qfun} is symmetric at state-action pairs $(x_{k},a_{k})$ and $(x^{\prime}_{k},a^{\prime}_{k})$ when

\begin{equation}
\begin{aligned}
    Q^{h}(x_{k},a_{k}) = Q^{h}(x^{\prime}_{k},a^{\prime}_{k})
\end{aligned}
\end{equation}
\end{myDef}

\section{DDPG with Symmetric Data Augmentation}\label{section_ddpg_sda}
This section develops a symmetric data augmentation method based on assumptions made in Section \ref{Section_symmetric_dynamic_model}. The augmented samples are used in the framework of DDPG to improve sample efficiency. 

\subsection{Symmetric data augmentation}

According to Equations \ref{symxt}, \ref{symat}, \ref{symxt+1}, \ref{symr}, The augmented samples can be obtained by mirroring the explored samples as

\begin{equation}\label{symmetricdataaugmenation}
    s^{\prime}_{k} = A s_{k} + B x^{*}
\end{equation}\\
where $s_{k}=[x_{k},a_{k},x_{k+1},r_{k}]^{T}$ is an explored sample, $s^{\prime}_{k}=[x^{\prime}_{k},a^{\prime}_{k},x^{\prime}_{k+1},r^{\prime}_{k}]^{T}$ is an augmented sample. The matrices $A$ and $B$ are defined as
\begin{equation}
    A = \begin{bmatrix}
    -1&0&0&0\\
    0&-1&0&0\\
    0&0&-1&0\\
    0&0&0&-1
    \end{bmatrix}, B= \begin{bmatrix}
    2\\
    0\\
    2\\
    0
    \end{bmatrix}
\end{equation}

\subsection{DDPG with symmetric data augmentation}

DDPG is a $Q$-learning–based approach that operates using minibatch updates. The critic is trained to approximate the $Q$-function, while the actor is trained to approximate the sub-optimal policy \cite{Lillicrap2016ICLR00}. This can be implemented through approximate value iteration (AVI). A replay buffer $\mathcal{D}$ is used to store explored samples and augmented samples:

\begin{equation}
    s_{k}\in \mathcal{D}, s^{\prime}_{k}\in \mathcal{D}
\end{equation}

In this way, the dataset is enriched. A minibatch consisting of both explored and augmented samples can be selected to train the critic and actor. This leads to an improved DDPG algorithm that exploits the augmented samples, namely the Deep Deterministic Policy Gradient with Symmetric Data Augmentation (DDPG-SDA). The pseudocode code is provided in Algorithm \ref{Algorithm_DDPG_SDA}.

\section{DDPG with Symmetric Critic Augmentation}\label{section_ddpg_sca}

This section investigates how to improve the sample utilization efficiency of DDPG-SDA. We show that using mixed samples within a minbatch is not effective for fully exploiting the augmented samples. Based on this analysis, we propose two modifications to the algorithmic structure: (1) $Q$-function approximation with two critic networks; (2) the two-step approximate value iteration. 

\subsection{Drawback of DDPG-SDA}

Mixing explored and augmented samples within a minibatch may not significantly improve the convergence of a policy, as the number of explored samples is reduced compared to a minibatch containing only explored data. One possible solution to this issue is to increase the minibatch size, but this may conversely degrade learning performance \cite{Obando2023}. Therefore, it is worthwhile to investigate ways to improve policy convergence without changing the size of minibatch.

\subsection{Dual critics for Q function approximation}

To exploit the augmented dataset more efficiently, we propose separate storage and sampling of explored and augmented samples. To this end, two replay buffers, denoted by $\mathcal{D}_{1}$ and $\mathcal{D}_{2}$, are used to store the explored and augmented samples separately:

\begin{equation}
    s_{k}\in \mathcal{D}_{1}, s^{\prime}_{k}\in \mathcal{D}_{2}
\end{equation}

As such, the two types of samples can be sampled separately and used to train two critics. Consequently, the approximation of the $Q$ function is through

\begin{equation}\label{piecewiseQ}
    Q^{h}(x_{k},a_{k})\approx \begin{cases}
        \hat{Q}^{h}_{\psi_{1}}(x_{k},a_{k}), & \text{if}\ (x_{k},a_{k})\in \mathcal{D}_{1}\\
        \hat{Q}^{h}_{\psi_{2}}(x_{k},a_{k}), &\text{if}\ (x_{k},a_{k})\in \mathcal{D}_{2}
    \end{cases}
\end{equation}\\
where $\psi_{1}$,$\psi_{2}$ are parameter sets of the two critics.

\subsection{Two-step approximate value iteration}

The two critics and the actor are trained based on a two-step approximate value iteration method: \\
\textbf{Step 1} (AVI on a minibatch from $\mathcal{D}_{1}$):
\begin{equation}\label{step1_PE}
    (\hat{Q}^{h}_{\psi_{1}})^{i+1}(x_{k},a_{k}) = r(x_{k},a_{k}) + \gamma (\hat{Q}^{h}_{\psi_{1}})^{i}(x_{k+1},(\mu_{\vartheta})^{i}(x_{k+1}))
\end{equation}

\begin{equation}
\label{step1_PI}
(\mu_{\vartheta})^{i+1}(x_{k})
= \arg\max_{\vartheta}
(\hat{Q}^{h}_{\psi_{1}})^{i+1}(x_{k}, \mu_{\vartheta}(x_{k}))
\end{equation}\\
\textbf{Step 2} (AVI on a minibatch from $\mathcal{D}_{2}$):
\begin{equation}\label{step2_PE}
    (\hat{Q}^{h}_{\psi_{2}})^{i+1}(x_{k},a_{k}) = r(x_{k},a_{k}) + \gamma (\hat{Q}^{h}_{\psi_{2}})^{i}(x_{k+1},(\mu_{\vartheta})^{i+1}(x_{k+1}))
\end{equation}
\begin{equation}\label{step2_PI}
        (\mu_{\vartheta})^{i+2} (x_{k})= \arg\max\limits_{\vartheta}(\hat{Q}^{h}_{\psi_{2}})^{i+1}(x_{k},\mu_{\vartheta}(x_{k}))
\end{equation}

\begin{myRem}
Step 1 trains the critic and actor networks on a minibatch of samples from $\mathcal{D}_{1}$, while Step 2 trains them again on a minibatch from $\mathcal{D}_{2}$. As a result, the critic networks are trained separately with explored and augmented samples, whereas the actor is trained on samples from both datasets. The corresponding parameter update equations are provided in \ref{appendix_API_gradient}.
\end{myRem}

\subsection{DDPG with symmetric critic augmentation} 

The proposed modifications make DDPG with symmetric critic augmentation (SCA). The advantage lies in the more effective utilization of augmented samples without increasing the minibatch size, which can potentially lead to faster policy convergence during learning (as demonstrated in the simulation results). The pseudocode is seen in Algorithm \ref{Algorithm_DDPG_SCA}.


\begin{algorithm}[H]
   \DontPrintSemicolon
   \KwData{Critic/actor learning rate $\eta_{\text{critic}},\eta_{\text{actor}}$, delay factor $\tau$, batch size $N$, discount factor $\gamma$, Ornstein–Uhlenbeck (OU) process parameters $\sigma,\theta,dt$ \cite{Doob1942}, maximal environment step $k_{\text{max}}$, maximal episode number $N_{\text{eps}}$}\;
   \textbf{Initialization:} Critic network parameter set $\psi$, actor network parameter set $\vartheta$, empty replay buffer $\mathcal{D}$, target critic network parameter set $\psi_{\text{target}}\leftarrow \psi$, target actor network parameter set $\vartheta_{\text{target}}\leftarrow \vartheta$\;
   \For{episode $\leftarrow$ 1 to $N_{eps}$}{
        \For{environment step $k \leftarrow 0$ to $k_{max} $}{
        Observe state $x_{k}$ and select an exploratory action \qquad \qquad \qquad \qquad \qquad \qquad \qquad \qquad \qquad \qquad \qquad \qquad \qquad \qquad \qquad \qquad \qquad \qquad \qquad $a_{k} = \mu_{\vartheta}(x_{k})+\epsilon$, $\epsilon \sim \text{OU}(\sigma,\theta,dt)$\;
        Execute $a_{k}$ to the environment\;
        Observe next state $x_{k+1}$ and reward $r_{k}$,         
        store a sample $s_{k}=(x_{k},a_{k},r_{k},x_{k+1})$ in $\mathcal{D}$\;
        Compute a symmetric sample: $s^{\prime}_{k}=As_{k}+Bx^{*}$, store $s^{\prime}_{k}$ in $\mathcal{D}$\;
            Randomly choose $N$ samples from $\mathcal{D}$ to compose a minibatch $\mathcal{B}$\;  
            Compute the target of $\mathcal{B}$:\qquad \qquad \qquad \qquad \qquad \qquad \qquad \qquad \qquad \qquad \qquad \qquad \qquad \qquad \qquad \qquad \qquad \qquad \qquad 
            $z^{i} = r^{i}_{k} +\gamma\hat{Q}^{h}_{\psi_{\text{target}}}[x^{i}_{k+1},\mu_{\vartheta_{\text{target}}}(x^{i}_{k+1})]$, $1\leq i \leq N$\;
            Update the critic: $\psi \leftarrow \psi - \eta_{\text{critic}}\nabla_{\psi}\frac{1}{N}\sum^{N}_{i=1}[\hat{Q}^{h}_{\psi}(x^{i}_{k},a^{i}_{k})-z^{i}]^{2}$\;
            Update the actor policy: $\vartheta \leftarrow \vartheta -\eta_{\text{actor}} \nabla_{\vartheta}\frac{1}{N}\sum^{N}_{i=1}\hat{Q}^{h}_{\psi}[x^{i}_{k},\mu_{\vartheta}(x^{i}_{k})]$\;
            Update the target networks:\qquad \qquad \qquad \qquad \qquad \qquad \qquad \qquad \qquad \qquad \qquad \qquad \qquad \qquad \qquad \qquad \qquad \qquad \qquad  $\psi_{\text{target}}\leftarrow \tau\psi_{\text{target}} + (1-\tau)\psi$, \qquad \qquad \qquad \qquad \qquad \qquad \qquad \qquad \qquad \qquad \qquad \qquad \qquad \qquad \qquad \qquad \qquad \qquad \qquad $\vartheta_{\text{target}} \leftarrow \tau\vartheta_{\text{target}}+(1-\tau)\vartheta$
            }
   }
   \caption{DDPG-SDA}\label{Algorithm_DDPG_SDA}
\end{algorithm}

\begin{algorithm}[H]
   \DontPrintSemicolon
   \KwData{Critic/actor learning rate $\eta_{\text{critic}},\eta_{\text{actor}}$, delay factor $\tau$, batch size $N$, discount factor $\gamma$, OU process parameters $\sigma,\theta,dt$, maximal environment step $k_{\text{max}}$, maximal episode number $N_{\text{eps}}$}\;
   \textbf{Initialization:} Critic network parameter set $\psi$, actor network parameter set $\vartheta$, empty replay buffer $\mathcal{D}$, target critic network parameter set $\psi_{\text{target}}\leftarrow \psi$, target actor network parameter set $\vartheta_{\text{target}}\leftarrow \vartheta$\;
   \For{episode $\leftarrow$ 1 to $N_{eps}$}{
        \For{environment step $k \leftarrow 0$ to $k_{max} $}{
        Observe state $x_{k}$ and select an exploratory action \qquad \qquad \qquad \qquad \qquad \qquad \qquad \qquad \qquad \qquad \qquad \qquad \qquad \qquad \qquad \qquad \qquad \qquad \qquad $a_{k} = \mu_{\vartheta}(x_{k})+\epsilon$, $\epsilon \sim \text{OU}(\sigma,\theta,dt)$\;
        Execute $a_{k}$ to the environment\;
        Observe next state $x_{k+1}$, reward $r_{k}$, store a sample $s_{k}=(x_{k},a_{k},r_{k},x_{k+1})$ in $\mathcal{D}_{1}$\;
        Compute a symmetric sample: $s^{\prime}_{k}=As_{k}+Bx^{*}$, store $s^{\prime}_{k}$ in $\mathcal{D}_{2}$\;
        Randomly choose $N$ samples from $\mathcal{D}_{1}$, $\mathcal{D}_{2}$ to compose minibatches $\mathcal{B}_{1}$, $\mathcal{B}_{2}$\;
        Compute the target of $\mathcal{B}_{1}$:\qquad \qquad \qquad \qquad \qquad \qquad \qquad \qquad \qquad \qquad \qquad \qquad \qquad \qquad \qquad \qquad \qquad \qquad \qquad 
        $y^{i} = r^{i}_{k} +\gamma\hat{Q}^{h}_{\psi_{1\text{target}}}[x^{i}_{k+1},\mu_{\vartheta_{\text{target}}}(x^{i}_{k+1})]$, $1\leq i \leq N$\;
        Update the critic: $\psi_{1} \leftarrow \psi_{1} - \eta_{\text{critic}}\nabla_{\psi_{1}}\frac{1}{N}\sum^{N}_{i=1}[\hat{Q}^{h}_{\psi_{1}}(x^{i}_{k},a^{i}_{k})-y^{i}]^{2}$\;
            Update the actor policy: $\vartheta \leftarrow \vartheta -\eta_{\text{actor}} \nabla_{\vartheta}\frac{1}{N}\sum^{N}_{i=1}\hat{Q}^{h}_{\psi_{1}}[x^{i}_{k},\mu_{\vartheta}(x^{i}_{k})]$\;
            Update the target networks:\qquad \qquad \qquad \qquad \qquad \qquad \qquad \qquad \qquad \qquad \qquad \qquad \qquad \qquad \qquad \qquad \qquad \qquad \qquad $\psi_{1\text{target}}\leftarrow \tau\psi_{1\text{target}} + (1-\tau)\psi$\qquad \qquad \qquad \qquad \qquad \qquad \qquad \qquad \qquad \qquad \qquad \qquad \qquad \qquad \qquad \qquad \qquad \qquad \qquad 
            $\vartheta_{\text{target}} \leftarrow \tau\vartheta_{\text{target}}+(1-\tau)\vartheta$\\
        Compute the target of $\mathcal{B}_{2}$:\qquad \qquad \qquad \qquad \qquad \qquad \qquad \qquad \qquad \qquad \qquad \qquad \qquad \qquad \qquad \qquad \qquad \qquad \qquad 
        $z^{i} = r^{i}_{k} +\gamma\hat{Q}^{h}_{\psi_{\text{2target}}}[x^{i}_{k+1},\mu_{\vartheta_{\text{target}}}(x^{i}_{k+1})]$, $1\leq i \leq N$\;
        Update the critic: $\psi_{2} \leftarrow \psi_{2} - \eta_{\text{critic}}\nabla_{\psi_{2}}\frac{1}{N}\sum^{N}_{i=1}[\hat{Q}^{h}_{\psi_{2}}(x^{i}_{k},a^{i}_{k})-z^{i}]^{2}$\;
            Update the actor policy: $\vartheta \leftarrow \vartheta -\eta_{\text{actor}} \nabla_{\vartheta}\frac{1}{N}\sum^{N}_{i=1}\hat{Q}^{h}_{\psi_{2}}[x^{i}_{k},\mu_{\vartheta}(x^{i}_{k})]$\;
            Update the target networks:\qquad \qquad \qquad \qquad \qquad \qquad \qquad \qquad \qquad \qquad \qquad \qquad \qquad \qquad \qquad \qquad \qquad \qquad \qquad  $\psi_{2\text{target}}\leftarrow \tau\psi_{2\text{target}} + (1-\tau)\psi$\qquad \qquad \qquad \qquad \qquad \qquad \qquad \qquad \qquad \qquad \qquad \qquad \qquad \qquad \qquad \qquad \qquad \qquad \qquad 
            $\vartheta_{\text{target}} \leftarrow \tau\vartheta_{\text{target}}+(1-\tau)\vartheta$
            }
   }
   \caption{DDPG-SCA}\label{Algorithm_DDPG_SCA}
\end{algorithm}

\section{Model} \label{section_model}

This section introduces the aircraft lateral dynamics. First, a continuous-time dynamical model is introduced. Next, this model is discretized using the Euler's method. Finally, the symmetry of discrete-time model is analyzed based on Theorem \ref{Theorem_symmetry}.

\subsection{Aircraft model}
\subsubsection{Dynamical model}

The aircraft dynamical model is described by the differential equations \cite{Ohta1979}:

\begin{equation}\label{aircraftmodel}
\begin{aligned}
     \dot{\phi} &= p\\
     \dot{p} &= L_{p}^{\prime}p + L_{r}^{\prime}r+L_{\beta}^{\prime}\beta + L_{\delta_{a}}^{\prime}\delta_{a}+L_{\delta_{r}}^{\prime}\delta_{r}\\    
     \dot{\beta}&=Y_{p}^{*}p + Y_{\phi}^{*}\phi + (Y_{r}^{*}-1)r + Y_{\beta}\beta + Y_{\delta_{a}}^{*}\delta_{a}+Y_{\delta_{r}}^{*}\delta_{r}\\ \dot{r}&=N_{p}^{\prime}p+N_{r}^{\prime}r+N_{\beta}^{\prime}\beta + N_{\delta_{a}}^{\prime}\delta_{a} +N_{\delta_{r}}^{\prime}\delta_{r}
\end{aligned}
\end{equation}\\
where $\phi$ is bank angle, $p$ is roll rate, $\beta$ is sideslip angle, $r$ is yaw rate. $\delta_{a}$ is the aileron deflection, $\delta_{r}$ is the rudder deflection. Aerodynamic coefficients are seen in Table \ref{Aircraft_Physical_Coefficients}.

\subsubsection{Discretization}

The differential equations \ref{aircraftmodel} can be discretized using the Euler's method \cite{Stetter1973} with a sampling step $T$:

\begin{equation}
\label{discreteaircraftmodel}
\begin{aligned}
\phi_{k+1} &= \phi_k + p_k T\\
p_{k+1} &= p_k + \big( L'_p p_k + L'_r r_k + L'_\beta \beta_k 
+ L'_{\delta_a} \delta_{a,k} + L'_{\delta_r} \delta_{r,k} \big) T \\
\beta_{k+1} &= \beta_k + \big( Y_p^* p_k + Y_\phi^* \phi_k 
+ (Y_r^* - 1) r_k + Y_\beta \beta_k 
+ Y_{\delta_a}^* \delta_{a,k} + Y_{\delta_r}^* \delta_{r,k} \big) T\\
r_{k+1} &= r_k + \big( N'_p p_k + N'_r r_k + N'_\beta \beta_k 
+ N'_{\delta_a} \delta_{a,k} + N'_{\delta_r} \delta_{r,k} \big) T
\end{aligned}
\end{equation}


Rewrite equations \ref{discreteaircraftmodel} as

\begin{equation}\label{dismodelstatespace}
    x_{k+1} = F(x_{k})x_{k} + G(x_{k}) u_{k}
\end{equation}\\
where $x_{k} = [\phi_{k},p_{k},\beta_{k},r_{k}]^{T}$, $u_{k} = [\delta_{a, k},\delta_{r, k}]^{T}$, and $F(x_{k}),G(x_{k})$ are given by

{\footnotesize
\begin{equation}
    F(x_{k})=\left[\begin{array}{cccc}
    1 & T & 0 & 0\\
    0 & 1+L^{\prime}_{p}T & L^{\prime}_{\beta}T & L^{\prime}_{r}T  \\
    Y_{\phi}^{*}T & Y_{p}^{*} T& 1+Y_{\beta}T & (Y^{*}_{r}-1)T\\
    0  & N_{p}^{\prime}T & N_{\beta}^{\prime} T & 1+N_{r}^{\prime}T
\end{array}\right], G(x_{k})=\left[\begin{array}{cc}
    0 & 0  \\
    L^{\prime}_{\delta a}T & L^{\prime}_{\delta r}T  \\
    Y_{\delta_{a}}^{*}\delta_{a,k}T & Y_{\delta_{r}}^{*}T\\ 
    N_{\delta_{a}}^{\prime}\delta_{a,k}T  & N_{\delta_{r}}^{\prime}\delta_{r,k}T
\end{array}\right]
\end{equation} 
}

\subsection{Symmetry analysis}

This subsection analyzes the symmetry of the discrete-time model \ref{dismodelstatespace}. Theorem \ref{Theorem_symmetry} classifies symmetry planes into two cases: $x^{*}=0$ and $x^{*}\neq 0$. The second case requires a stricter constraint on the system matrix $F(x_{k})$. Equations \ref{discreteaircraftmodel} also exhibit coupling effects between roll and yaw channels, which are captured in $F(x_{k})$. The coupling effects would influence symmetry of the overall system.


Apply the assumptions of Theorem \ref{Theorem_symmetry} to system \ref{dismodelstatespace}:

\begin{equation}
\begin{aligned}
   &\frac{x_{k}+x^{\prime}_{k}}{2} = x^{*}\\
    &a_{k} = -a^{\prime}_{k}
\end{aligned}
\end{equation}\\
where $x_{k} = [\phi_{k},p_{k},\beta_{k},r_{k}], x^{\prime}_{k} =[\phi^{\prime}_{k},p^{\prime}_{k},\beta^{\prime}_{k},r^{\prime}_{k}]$ are states symmetric about the reference plane $x^{*}=[\phi^{*}_{k},p^{*}_{k},\beta^{*}_{k},r^{*}_{k}]$, $a_{k}=[\delta_{a,k},\delta_{r,k}],a^{\prime}_{k}=[\delta^{\prime}_{a,k},\delta^{\prime}_{r,k}]$ are symmetric action pairs about $0\in\mathbb{R}^{m}$. 

Because the conditions $F(x_{k})=F(x^{\prime}_{k})\neq 0 \in\mathbb{R}^{n\times n}, G(x_{k})=G(x^{\prime}_{k})$ hold for system \ref{dismodelstatespace}, one can conclude from case (1) of Theorem \ref{Theorem_symmetry} that

\begin{equation}
\begin{aligned}
   \frac{x_{k+1}+x^{\prime}_{k+1}}{2} = 0 \in\mathbb{R}^{n}\\
\end{aligned}
\end{equation}\\
where $x_{k+1} = [\phi_{k+1},p_{k+1},\beta_{k+1},r_{k+1}]$, $x^{\prime}_{k+1} =[\phi^{\prime}_{k+1},p^{\prime}_{k+1},\beta^{\prime}_{k+1},r^{\prime}_{k+1}]$.

\begin{table}[h]
\caption{Aerodynamic coefficients (adopted from \cite{Ohta1979})}
\label{Aircraft_Physical_Coefficients}
\begin{center}
\begin{tabular}{c|l|c|l}
\hline
Parameter & Value & Parameter & Value\\
\hline
$L^{\prime}_{p}$ & -1.699 & $Y^{*}_{p}$      & 0  \\
$L^{\prime}_{r}$ &  0.172 & $Y^{*}_{\phi}$   & 0.0488 \\
$L^{\prime}_{\beta}$ & -4.546 & $Y^{*}_{r}$  & 0\\
$N^{\prime}_{p}$ & -0.0654 & $Y_{\beta}$     & -0.0829\\
$N^{\prime}_{r}$ & -0.0893 & $L^{\prime}_{\delta_{a}}$ & 27.276\\
$N^{\prime}_{\beta}$ & 3.382&$L^{\prime}_{\delta_{r}}$ & 0.576\\
$Y^{*}_{\delta_{a}}$ & 0 & $Y^{*}_{\delta_{r}}$ & 0.116\\
$N_{\delta_{a}}$ & 0.395 & $N_{\delta_{r}}$ &-1.362\\
\hline
\end{tabular}
\end{center}
\end{table}

\section{Conditioning for Action Policy Smoothness} \label{section_action_smoothness}


Conditioning for Action Policy Smoothness (CAPS) techniques have been developed to achieve smooth control of a quadrotor in offline RL frameworks \cite{Mysore2021ICRA00}, and adapted for smooth control of the Flying-V and Cessna Citation PH-LAB aircraft \cite{Gavra202447,Dally2022AIAA00,Vieira2025AIAA00,Homola2025AIAA00}. The core idea is to incorporate smoothness losses into policy optimization, which can be viewed as a multiple-objective optimization problem that explicitly considers action smoothness. This approach encourages the actor to produce similar actions for neighboring input states that vary across spatial or temporal scales.

For a minibatch, the spatial smoothness loss can be formulated as

\begin{equation}
    L_{s} = \frac{1}{N_{s}}\sum^{N_{s}}_{j=1}\left\Vert \mu_{\vartheta}(\overline{x}^{j}_{k})-\mu_{\vartheta}(x^{j}_{k})\right\Vert_{2}, \quad j=1,2,\cdots,N_{s}
\end{equation}\\
where $x_{k}$ is the input of the actor at time step $t$, $\overline{x}_{k}$ is the biased $x_{k}$ in spatial scale, $j$ is the index of $N_{s}$ samples from the distribution of $x_{k}$. The spatial smoothness loss penalizes changes in actions with respect to biased states caused by measurement noise, which helps improve the generalization and robustness of the policy.


For a minibatch, the temporal smoothness loss can be formulated as

\begin{equation}
    L_{t} = \frac{1}{N}\sum^{N}_{i=1}\left\Vert \mu_{\vartheta}(x^{i}_{k+1})-\mu_{\vartheta}(x^{i}_{k})\right\Vert_{2}, \quad i=1,2,\cdots,N\end{equation}\\
where $x_{k+1},x_{k}$ are the input of the actor at steps $k$ and $k+1$, $i$ is the index of $N$ samples in a minibatch.

The temporal smoothness loss measures changes in the actor’s output in response to changes in the input in temporal scale. The abrupt variations in actions between consecutive time steps are penalized so that the actor learns to produce a series of slow-varying actions over time.

A multi-objective optimization is then formulated as

\begin{equation}
    h(x^{i}_{k}) = \arg\max_{h}\left(\sum^{n}_{i=1}\left[\hat{Q}^{h}_{\psi}\left(x^{i}_{k},h(x^{i}_{k})\right)\right]-\lambda_{1}L_{s}-\lambda_{2}L_{t}\right)
\end{equation}\\
where the first term measures the long-term state–action value, and $\lambda_{1}\geq 0,\lambda_{2}\geq 0$ are weights for the smoothness losses. These parameters can be tuned to balance policy performance in maximizing the estimated state–action value and ensuring action smoothness.
\section{Simulation}\label{section_simulation}

This section presents the flight control simulation results. The symmetry-informed RL algorithms are evaluated during both the online training and online operation phases. Furthermore, the generalization capability of the trained actors is assessed by tracking a bank-angle reference that was not used during policy training.

\subsection{Environment settings}
The environment is set as the aircraft model \ref{aircraftmodel}. These differential equations are integrated using the fourth-order Runge–Kutta method with a time step of 0.1s. Each episode consists of 300 time steps. The weights of the critic, actor are initialized using the Kaiming distribution \cite{He2015ICCV00} provided by PyTorch’s linear module. Sampling from replay buffers is performed without replacement, to ensure the full diversity of the samples in a minibatch, improving sample efficiency in each update of networks. The aircraft states are initialized randomly by uniform distributions, i.e. $\phi_{0}\sim \mathcal{U}(0^{\circ},20^{\circ})$, $p_{0}\sim \mathcal{U}(0^{\circ}/\text{s},10^{\circ}/\text{s})$, $\beta_{0}\sim\mathcal{U}(0^{\circ},20^{\circ})$, $r_{0}\sim \mathcal{U}(0^{\circ}/\text{s},10^{\circ}/\text{s})$. Actuator actions are constrained as $\delta_{a},\delta_{r}\in[-57.3^{\circ}, 57.3^{\circ}]$, which defines an action space $\mathcal{A}=[-57.3^{\circ},57.3^{\circ}]\times[-57.3^{\circ},57.3^{\circ}]\in\mathbb{R}^{2}$. The bank angle reference $\phi_{\text{ref}}$ is set as a square-wave signal with a period of $T=3\text{s}$. The amplitude is selected at the beginning of each episode by sampling it from a uniform distribution $\mathcal{U}(0^{\circ},20^{\circ})$. The environment state is augmented with tracking error as $[e_{\phi},\phi,p,\beta,r]^{T}\in\mathbb{R}^{5}$, where $e_{\phi}=\phi-\phi_{\text{ref}}$. The reward function is shaped to promote bank angle tracking and sideslip angle stabilization, i.e. $\phi\rightarrow \phi_{\text{ref}},\beta\rightarrow 0$. The aggressive control behaviors are constrained by penalizes excessive angular rates and control efforts. The reward function is given as 



\begin{equation}
\begin{aligned}  \label{rewardfunction} 
    r_{k} =- 10\Vert\text{clip}(5e_{k},-1,1)\Vert_{1}-\Vert p_{k}\Vert_{1}-\Vert r_{k}\Vert_{1}  -0.01\Vert\delta_{a,k}\Vert_{1}-0.01\Vert\delta_{r,k}\Vert_{1}
\end{aligned}
\end{equation}\\
where the tracking error vector is defined as $e_{k}=[e_{\phi,k},e_{\beta,k}]^{T}\in\mathbb{R}^{2\times1},e_{\phi,k}=\phi_{k}-\phi_{\text{ref},k},e_{\beta,k}=\beta_{k}-0$. The function $\text{clip}(\cdot)$ is used to amplify tracking errors $e_{\phi,k}$ in the set $\{e_{\phi,k}|0.2<\vert e_{\phi,k}\vert<1\}$, and $e_{\beta,k}$ in the set $\{e_{\beta,k}|0.2<\vert e_{\beta,k}\vert<1\}$.  $\Vert\cdot\Vert_{1}$ is the $1$-norm operator. Figure \ref{Figure_brief_architecture_of_RL_based_flight_control_system} illustrates the overall architecture of the RL-based flight control system. Figure \ref{bank_angle_references} shows the bank angle references. The asymmetric references help to test the agent's ability of ’imagination’ for state space, instead of using exploration to cover the state space.

The critic network is a multi-layer-perception (MLP) with the input vector $[e_{\phi},\phi,p,\delta_{a},\beta,r,\delta_{r}]^{T}\in\mathbb{R}^{7}$ and the output of the estimated state-action value. An activation function $-\text{abs}(\cdot)$ is applied in the output layer to ensure the negative definiteness of the estimated state-action value function. The actor network is a multi-layer-perception (MLP) with the input vector $[e_{\phi},\phi,p,\beta,r]^{T}\in\mathbb{R}^{5}$ and the output of the control surface deflections $[\delta_{a}, \delta_{r}]^{T}\in\mathbb{R}^{2}$. A scaled $\tanh(\cdot)$ activation function is applied in the output layer to constrain the actions within the actuator limits. The target critic and actor share the same architectures as their corresponding primary networks.

The overall training process consists of five independent instances, each initialized with a different random seed. Each instance runs for 3000 episodes. The baseline algorithms include: (1) DDPG, used to evaluate the effectiveness of augmented samples compared with DDPG-SDA; and (2) DDPG with two updates per iteration, denoted as DDPG (2 updates), used to compare with DDPG-SCA in terms of the effectiveness of augmented samples. The hyperparameters are listed in Table \ref{RL_algorithm_hyperparameters}.



\begin{myRem}
    As the dataset used for off-policy training is collected by an exploration policy, the learned policy may propose actions far from those in the dataset. In such cases, the $Q$-value estimates can become highly unreliable, leading to inaccurate policy improvement \cite{vanHasselt2010}. This problem is especially pronounced when tracking errors change in a large range during training, as the policy has not yet learned to follow the reference. To address this, tracking errors can be clipped to ensure that updates to the critic and actor remain within a safe range, thereby reducing reliance on Out-of-Distribution (OOD) actions \cite{An2021}. Using $\text{clip}(\cdot)$ prevents large TD errors and $Q$-value updates caused by excessive tracking errors, which could otherwise destabilize the updates of the critic and actor.
\end{myRem}


\begin{myRem}
    Regularizing the angular rates in the reward design helps reduce overshoot and the amplitude of oscillations during tracking error convergence. This approach is consistent with the rate-feedback control design in the Linear Quadratic Regulator (LQR) \cite{Lewis2012} and quadratic-cost Approximate Dynamic Programming (ADP) \cite{Heydari2016}, where the rate-weighting parameters in the state-weighting matrix (commonly denoted as $Q$) are tuned to achieve similar effects.
\end{myRem}


\begin{table}[h]
\caption{RL Algorithm Hyperparameters}
\label{RL_algorithm_hyperparameters}
\begin{center}
\begin{tabular}{l|l}
\hline
  Parameter & Value\\
\hline
 Critic learning rate $\eta_{\text{critic}}$ & 0.001\\
 Actor learning rate $\eta_{\text{actor}}$ & 0.001\\
 Delay factor $\tau$ & 0.01\\
Discount factor $\gamma$& 0.99\\
OU process parameters $\sigma, \theta, dt$ & 0.015,0.1,0.01\\
Buffer size & 9$\times$10$^{6}$\\
 minibatch size $N$&256\\
 Optimizer & Adam\\
 Hidden layer structure & 64$\times$64\\
 Critic activation functions & ReLU\\
 Actor output-layer activation function & tanh\\
Actor hidden-layers activation function & tanh-ReLU\\
Spatial smoothness loss weight $\lambda_{1}$ & 3.5$\times$10$^{-5}$\\
Temporal smoothness loss weight $\lambda_{2}$ & 1.11$\times$10$^{-5}$\\
\hline
\end{tabular}
\end{center}
\end{table}

\begin{minipage}[t]{0.63\textwidth}
    \centering
    \includegraphics[width=\linewidth]{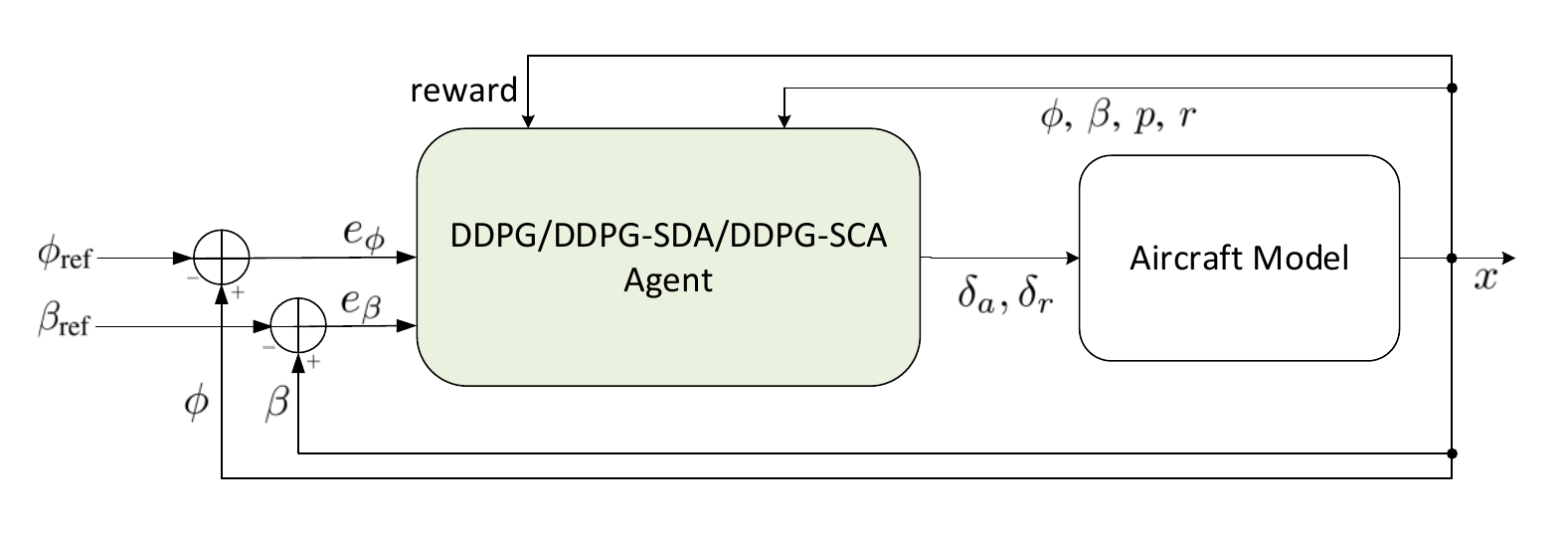}
    \captionof{figure}{The architecture of the RL-based flight control system.}\label{Figure_brief_architecture_of_RL_based_flight_control_system}
\end{minipage}
\hspace{1mm}
\begin{minipage}[t]{0.33\textwidth} 
    \centering
    \includegraphics[width=\linewidth]{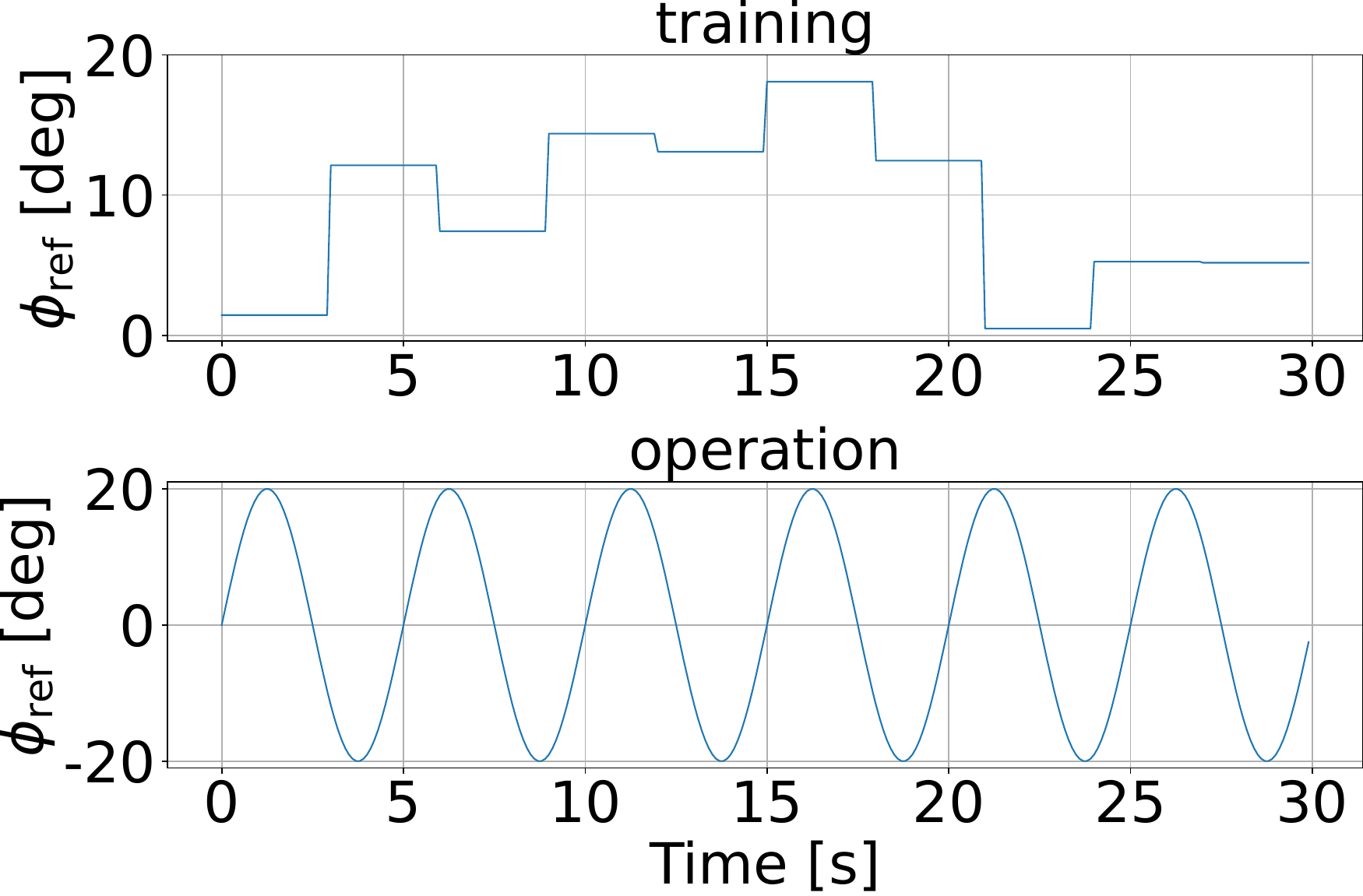} 
    \captionof{figure}{Bank angle references during training and operation. The training reference varies randomly in amplitude within $[0,20^{\circ}]$ with each value held for 3s.}\label{bank_angle_references}
\end{minipage}



\subsection{Training results}





\begin{figure}
\centering 
\includegraphics[width=0.65\linewidth]
{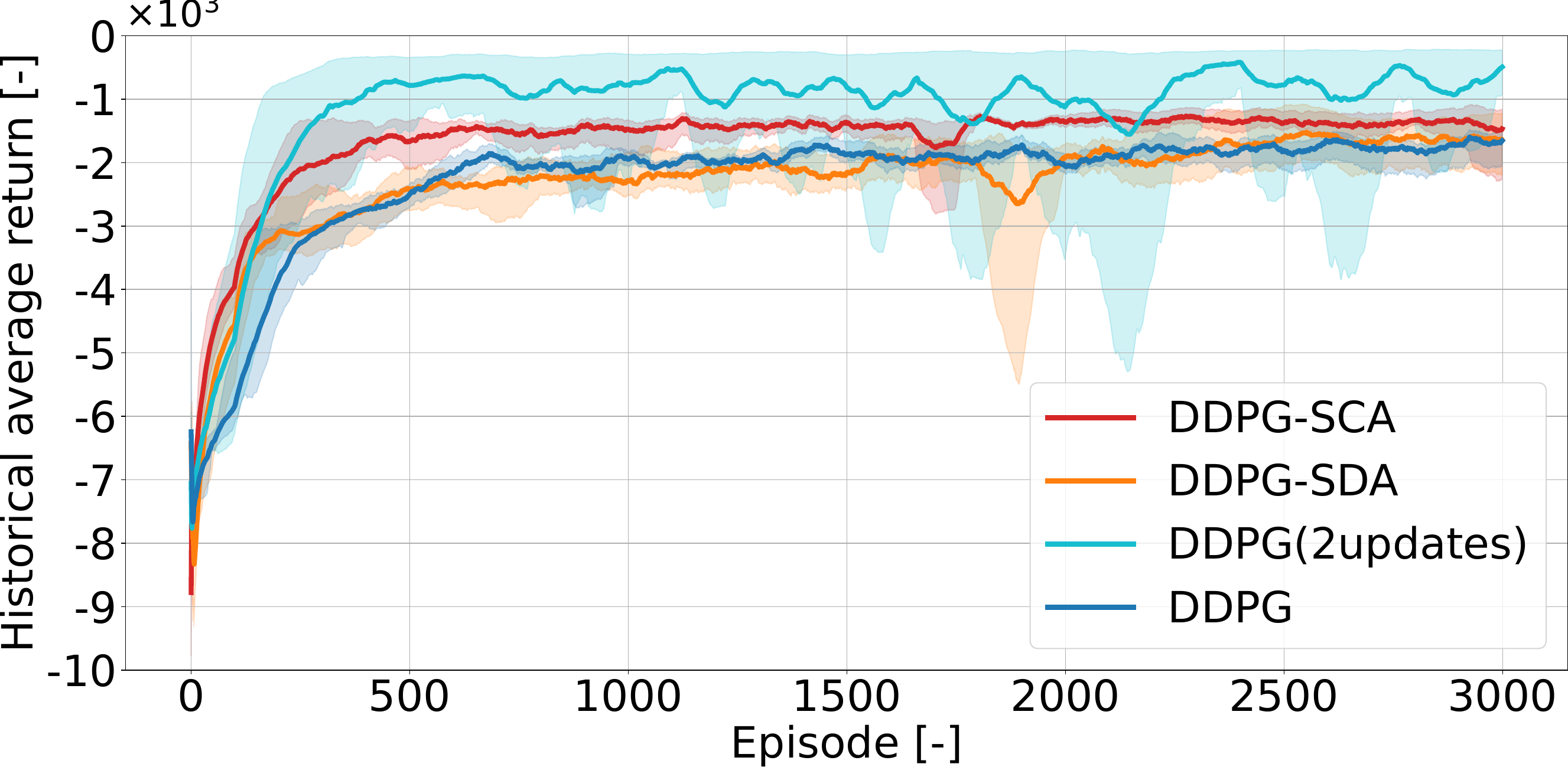}
\caption{Online training performance over 3000 episodes.
The solid line represents the mean of average returns across five independent runs, while the dashed lines indicate the maximum and minimum values.} \label{learning_performance_3000_episodes}
\end{figure}

Figure \ref{learning_performance_3000_episodes} presents the training performance over 3000 episodes. Table \ref{Training_performance_3000} summarizes the statistics results. The rolling average return is defined as the mean return over the most recent 100 episodes, providing a smoothed performance measure. All algorithms have converged to comparable episode returns by episode 3000. DDPG (2 updates) achieves a higher rolling average return than DDPG, which demonstrates that a twice-update frequency reaches a higher final value than a once-update frequency. This advantage of the training setup also applies to symmetry-informed algorithms if the actors are trained twice per iteration on the explored dataset. In Figure \ref{learning_performance_3000_episodes}, DDPG-SCA updates the actor twice per iteration, but one of the updates uses augmented samples from other regions of the state space, which contribute less to the rolling average return in the training scenarios. The average rate quantifies the rate of change of the rolling average return. DDPG-SDA exhibits a higher rolling average rate than DDPG during the first 500 episodes, enabling faster policy convergence. This verifies the improved performance of mixing explored samples and symmetric samples compared to using only symmetric samples. However, their performance becomes comparable after sufficient exploration. Another advantage of DDPG-SDA over DDPG lies in the fact that it requires fewer interactions with the environment to fill the replay buffer and therefore reduces the costs of exploration. The higher average rate of DDPG-SCA (14.212) compared to DDPG-SDA (8.045) indicates that the two-step approximate value iteration outperforms the one-step method in terms of accelerating policy convergence, due to increased update frequency of the actor. The low average rates during episodes 2500–3000 confirm that the policies have been sufficiently learned.

\begin{table*}
\footnotesize
\caption{Training performance}
\label{Training_performance_3000}
\begin{center}
\resizebox{.98\columnwidth}{!}{
\begin{tabular}{l|l|l|l|l}
\hline
Metric/Algorithm & DDPG-SCA & DDPG-SDA & DDPG& DDPG (2updates)\\
\hline
Rolling average return at episode 500 &-1672.669 $\pm$ 250.239 &-2408.224 $\pm$ 234.935 & -2499.168 $\pm$ 147.257& -784.213$\pm$435.842\\
Average rate during episodes 1-500&14.212$\pm$2.118 &8.045$\pm$1.058 & 7.489$\pm$2.805& 12.642$\pm$1.918\\
\hline
Rolling average return at episode 3000 &-1468.781 $\pm$ 395.151 & -1643.008 $\pm$ 345.124 & -1654.506 $\pm$ 197.981&-492.559$\pm$262.708\\
Average rate during episodes 2500-3000 & -0.197$\pm$0.678 & -0.056$\pm$0.212   & 0.326$\pm$0.099&-0.197$\pm$0.678\\
\hline
\end{tabular}
}
\end{center}
\end{table*}

\subsection{State space coverage}

\begin{figure}[h]
    \centering

    \begin{minipage}{1.0\textwidth}
        \centering
        \includegraphics[width=0.9\textwidth]{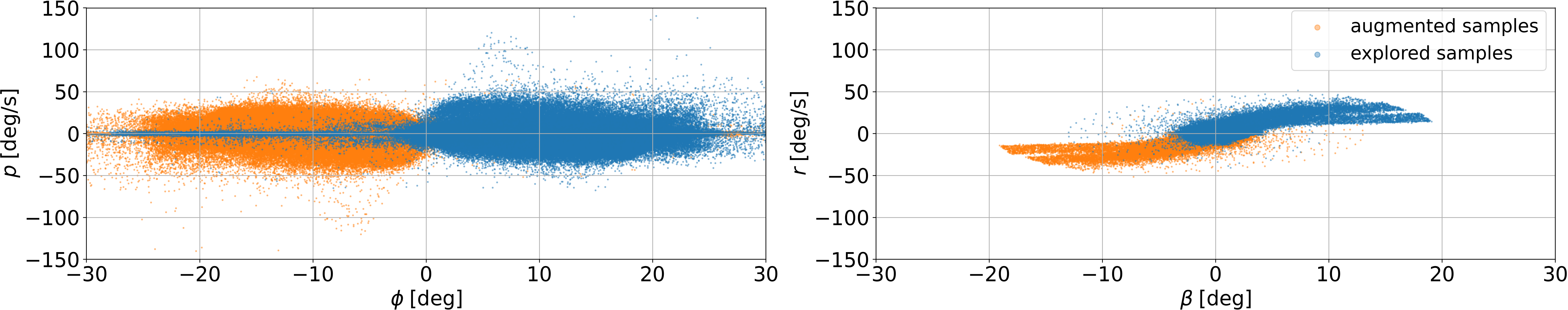}
        
        (a) Asymmetric distribution of initial states and reference ($\sigma=0.015$).
    \end{minipage}


    \begin{minipage}{1.0\textwidth}
        \centering
        \includegraphics[width=0.9\textwidth]{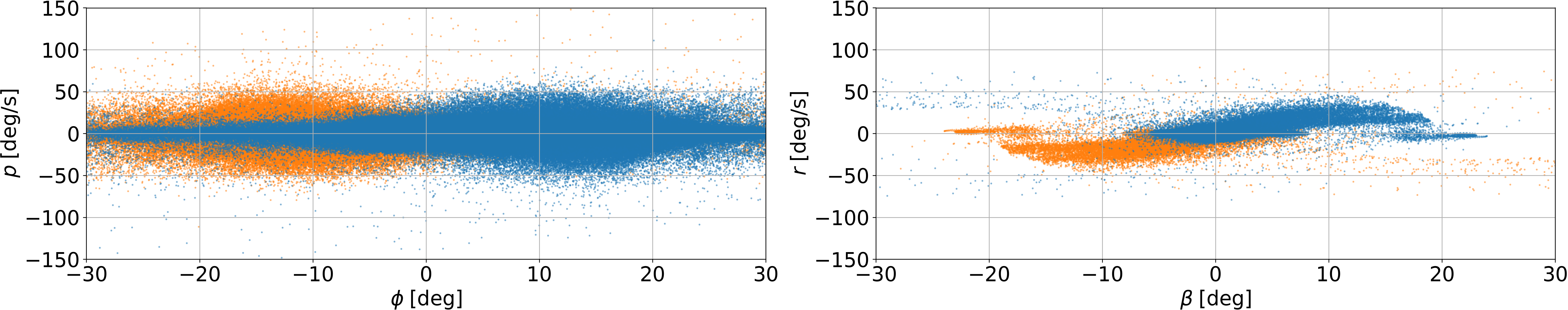}
        
        (b) Asymmetric distribution of initial states and reference ($\sigma=0.045$).
    \end{minipage}


    \begin{minipage}{1.0\textwidth}
        \centering
        \includegraphics[width=0.9\textwidth]{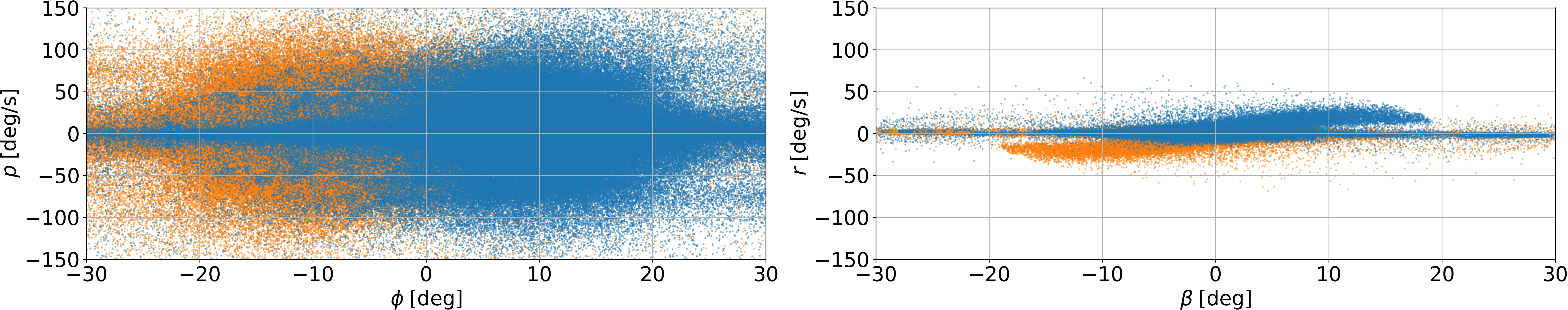}
        
        (c) Asymmetric distribution of initial states and reference ($\sigma=0.075$).
    \end{minipage}


    \begin{minipage}{1.0\textwidth}
        \centering
        \includegraphics[width=0.9\textwidth]{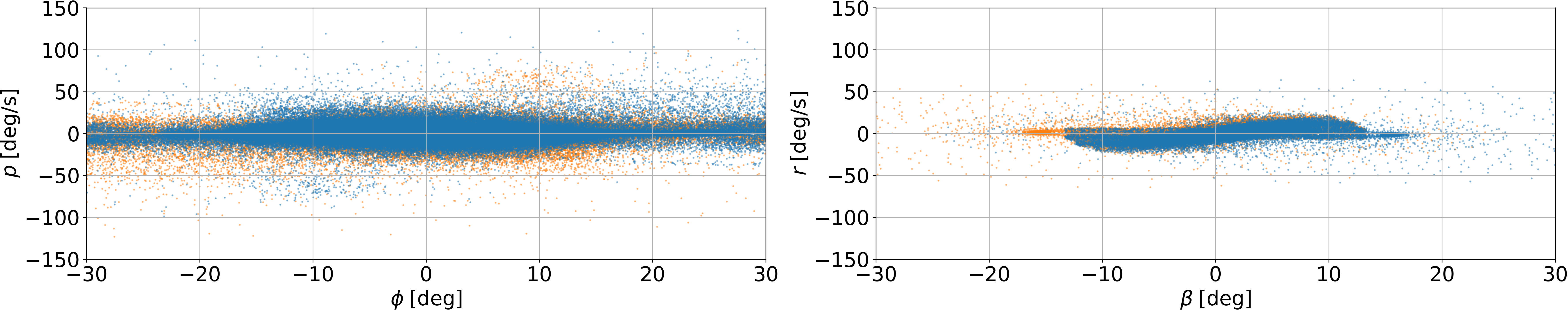}
        
        (d) Symmetric distribution of initial states and reference ($\sigma=0.015$).
    \end{minipage}

    \caption{Distribution of one-instance training samples in the local state space, projected onto two two-dimensional planes.}\label{state_space}
\end{figure}

Figure \ref{state_space} shows the training samples in a local state space $\mathcal{S}_{\text{local}} =\big\{ [\phi,p,\beta,r]^{T}\in\mathbb{R}^{4}|\phi\in[-30^{\circ},30^{\circ}],p\in[-150^{\circ}/\mathrm{s},150^{\circ}/\mathrm{s}],\beta\in[-30^{\circ},30^{\circ}],r\in[-150^{\circ}/\mathrm{s},150^{\circ}/\mathrm{s}]\big\}$. The symmetric distribution of explored and augmented samples with respect to the origin reflects the inherent symmetry of the data augmentation process. The OU noise is a random noise added along the actor’s policy, so that the exploration performance depends on the actor's policy and the randomness of OU noise. The explored samples cover only a local region of $\mathcal{S}_{\text{local}}$ due to the limited capability of the exploration policy. However, the augmented samples compensate for the corresponding symmetric regions that lack exploration, primarily within $[\phi,p,\beta,r]\in[-30^{\circ},0^{\circ}]\times[-50^{\circ}/\mathrm{s},50^{\circ}/\mathrm{s}]\times[-20^{\circ},10^{\circ}]\times[-50^{\circ}/\mathrm{s},0^{\circ}/\mathrm{s}]$. Discretize $\mathcal{S}_{\text{local}}$ with a resolution of 1$^{\circ}$ along the $\phi$ and $\beta$ axes, and 10$^{\circ}$/s along the $p$ and $r$ axes. The coverage rate is defined as the percentage of four-dimensional unit grids occupied by at least one sample. As shown in Table \ref{coverage_rate_statistics} and Figures \ref{state_space} (a)(b)(c), the exploration capability of OU noise, represented by the variances $\sigma = 0.015, 0.045, 0.075$, increases the coverage rate for $S_{\text{local}}$. The augmented samples still compensate for the corresponding symmetric regions with high angles and rates, where exploration becomes limited as the policy converges. The sample augmentation approach also saves the exploration effort. Therefore, symmetric data augmentation is effective for acquiring samples that may not be easily obtained through exploration and, at the same time, reduces exploration effort.

To be more illustrative, the symmetric distributions for the initial states are used: $\phi_{0}\sim \mathcal{U}(-20^{\circ},20^{\circ})$, $p_{0}\sim \mathcal{U}(-10^{\circ}/\text{s},10^{\circ}/\text{s})$, $\beta_{0}\sim \mathcal{U}(-20^{\circ},20^{\circ})$, and $r_{0}\sim \mathcal{U}(-10^{\circ}/\text{s},10^{\circ}/\text{s})$, along with a reference amplitude distribution $\mathcal{U}(-20^{\circ},20^{\circ})$. Figure~\ref{state_space} (d) shows that the explored samples achieve symmetric coverage of $S_{\text{local}}$ with a coverage rate of 0.652\%, which increases to 1.006\% when augmented samples are included. This indicates that the explored samples largely overlap with the augmented samples, suggesting that OU noise ($\sigma=0.015$) can generate samples that extensively cover $\mathcal{S}_{\text{local}}$, analogous to SDA. Moreover, a large number of exploration episodes ensures sufficiently dense sampling of $\mathcal{S}_{\text{local}}$.

\begin{table}[h]
\caption{Coverage rate of $\mathcal{S}_{\text{local}}$}
\label{coverage_rate_statistics}
\begin{center}
\resizebox{.98\columnwidth}{!}{
\begin{tabular}{l|l|l|l|l}
\hline
 Distributions of initial state and reference & Asymmetric ($\sigma=0.015$) &  Asymmetric ($\sigma=0.045$) &  Asymmetric ($\sigma=0.075$) & Symmetric ($\sigma=0.015$)\\
\hline
Explored samples & 0.286\% &  0.467\%  & 1.111\% & 0.652\%  \\
Explored \& Augmented samples &0.542\% &0.796\% & 1.865\% & 1.006\%\\
\hline
\end{tabular}
}
\end{center}
\end{table}





\subsection{Attitude tracking}

This subsection evaluates the attitude tracking performance under the controllers trained with RL approaches. The reference signals are given as $\phi_{\text{ref}}(t) = 20^{\circ}\sin(0.2\pi t), \beta_{\text{ref}}=0$. We assess how the trained agents perform with a bank angle reference that was not used during policy training. A notable characteristic of this bank angle reference is its range across both positive and negative regions, which poses a challenge to the generalization of actors trained only on the positive region provided by the reference during training. During the operation phase, the actor parameters are fixed to the values obtained after 3000 episodes of online training. The initial states are randomized according to the same uniform distributions used during the training phase.

\begin{figure}
\centering 
\includegraphics[width=1.0\linewidth]
{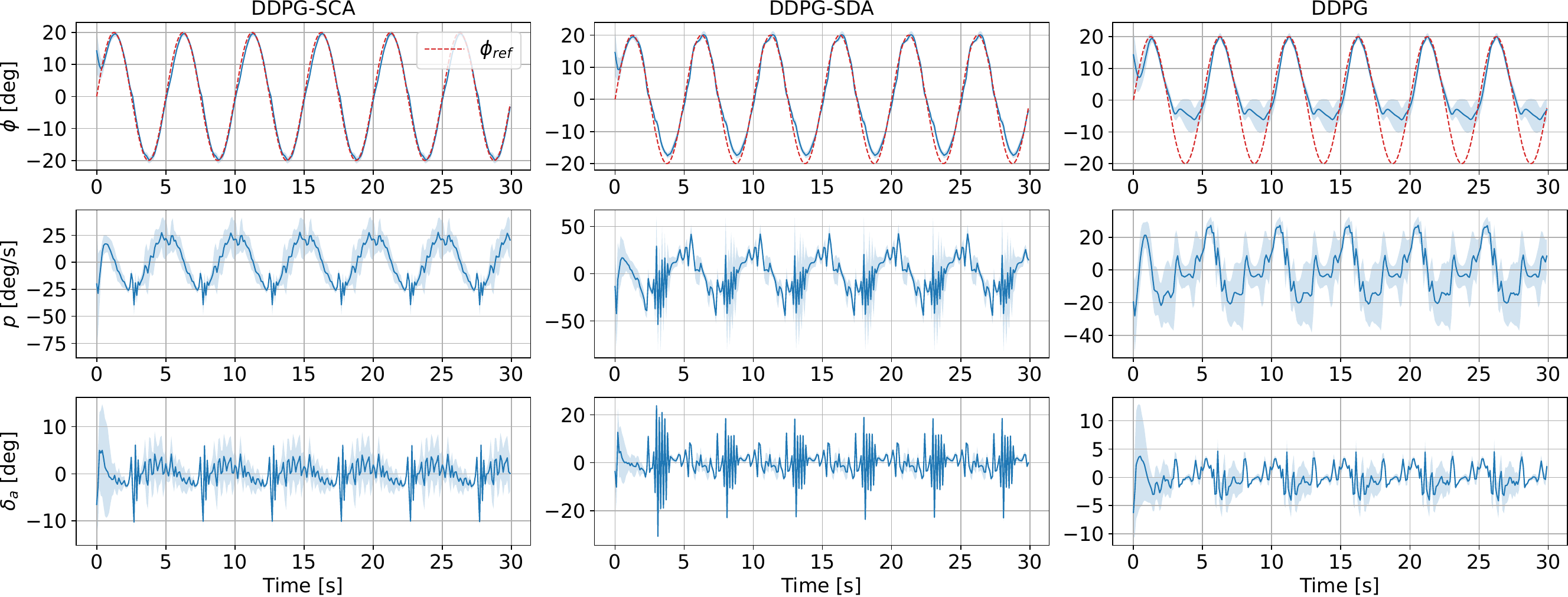}
\caption{States and controls in roll channel. The solid lines represent the mean across 5 instances, and the shaded regions indicate one standard deviation.}\label{figure_roll_channel_tracking}
\end{figure}

\begin{figure}
\centering 
\includegraphics[width=1.0\linewidth]
{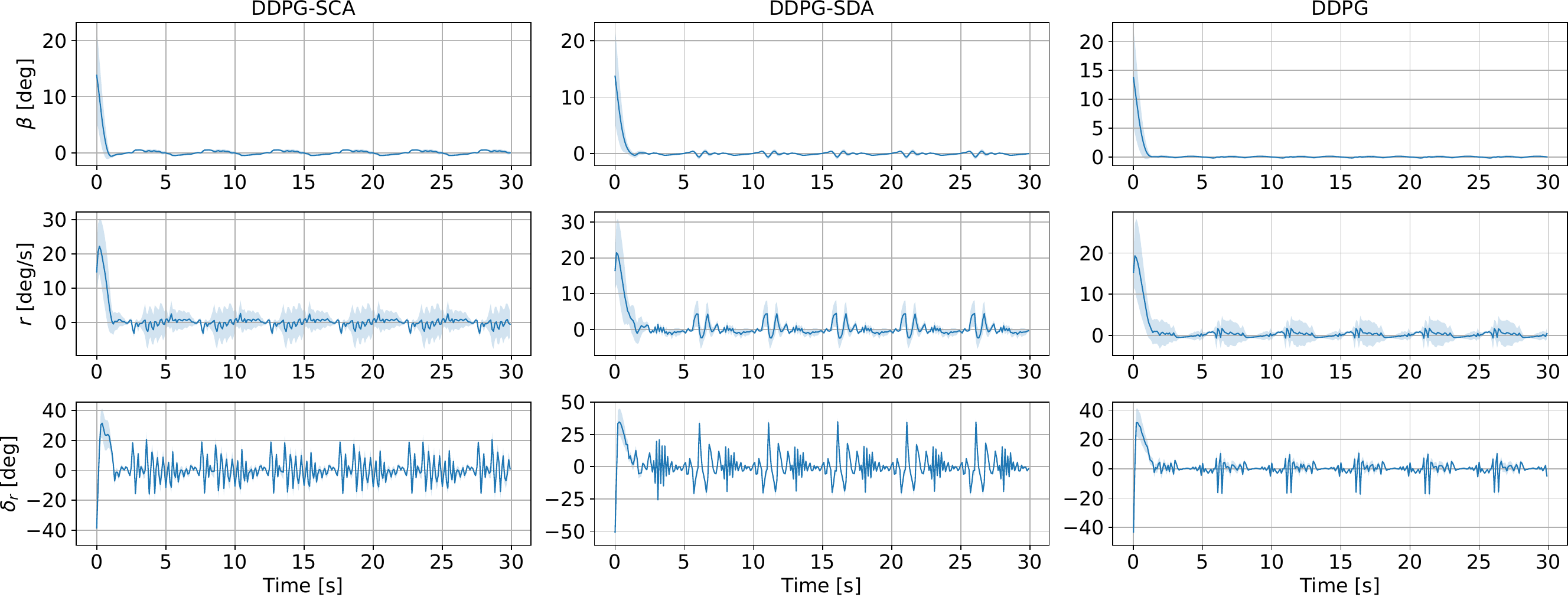}
\caption{State and control in yaw channel. The solid lines represent the means across 5 instances, and the shaded regions indicate one standard deviation.}\label{figure_yaw_channel_tracking}
\end{figure}

Figures \ref{figure_roll_channel_tracking} and \ref{figure_yaw_channel_tracking} exhibit the curves of states and actions. It can be observed that the controller trained with DDPG fails to track the bank angle reference in the negative part of bank angle reference, as no samples from this region are collected and used during policy training. Consequently, the controller must produce actions based solely on the neural network's generalization capability. In contrast, the symmetry-informed approaches exhibit comparable tracking performance in the positive and negative parts of the bank angle reference, owning to the additional samples generated by the symmetric data augmentation algorithm.

Subsequently, we evaluate the tracking performance quantitatively in terms of tracking accuracy and control effort. The first metric, Integral of the Absolute Error Mean (IAEM) \cite{Shinner1978,Guo2017}, evaluates tracking error $e_{k}$ over $n$ trajectories, each with a time horizon $k=0,1,\cdots,n_{e}$. It is computed as $\text{IAEM} =\frac{1}{n}\sum^{n}_{i=1}(\text{IAE})_{i} $. The auxiliary metric $(\text{IAE})_{i}=\sum^{n_{e}}_{k=0} \Vert e^{i}_{k}\Vert_{1} T$ evaluates the tracking error over a single trajectory indexed by $i$. Similar to the definition of IAEM, the second metric, the Integral of the Absolute Control Mean (IACM), evaluates the control effort over $n$ trajectories as $\text{IACM} = \frac{1}{n}\sum^{n}_{i=1}(\text{IAC})_{i}$, where the auxiliary metric $(\text{IAC})_{i}=\sum^{n_{e}}_{k=0} \Vert a^{i}_{k}\Vert_{1} T$ measures the integral of the absolute control input $a^{i}_{k}$ over steps $k=0,1,\cdots,n_{e}$ for the $i$-th trajectory. The results are presented in Table \ref{Attitude_tracking_results}. DDPG-SCA and DDPG-SDA achieve comparable $\phi$-tracking performance in the roll channel. In contrast, DDPG exhibits poor $\phi$-tracking performance due to the loss of symmetric samples. Therefore, the SDA algorithm improves the tracking performance of the actors in regions of the state space that lack exploration. All three methods achieve effective $\beta$-regulation, but periodic oscillations are observed in the yaw rate and the rudder deflection due to coupling effects from the roll channel.

However, if the exploration of the state space is already symmetric or sufficiently rich, SDA will not provide such significant improvement, as the policies are trained on datasets with similar coverage. The policies will therefore behave similarly. The advantages of SDA arise under the following conditions:
(1) asymmetric initial states and references, leading to asymmetric state trajectories, so the agent has no access to samples from symmetric trajectories;
(2) insufficient exploration, for example, when the variance of OU noise is small, causing the policy to converge quickly and explore a limited area of $\mathcal{S}$. The focus of SDA is on imagination of the agent for state space, instead of using exploration to cover the state space.

\begin{table}[h]
\caption{Statistical results of tracking control}
\label{Attitude_tracking_results}
\begin{center}
\begin{tabular}{l|l|l|l|l}
\hline
Channel & Metric & DDPG-SCA & DDPG-SDA & DDPG\\
\hline
Roll channel & IAEM &  1.044  & 1.136 & 5.225  \\
&IACM & 22.630 & 26.898 & 19.649\\
\hline
Yaw channel&IAEM &0.232 & 0.251 & 0.212\\
&IACM & 54.466  & 56.647 & 50.211\\
\hline
\end{tabular}
\end{center}
\end{table}

\section{Conclusion} \label{section_conclusion}

This paper investigates symmetry-informed RL algorithms that exploit the symmetry of dynamical models to improve sample efficiency and reduce exploration costs. The aircraft model exhibits symmetry properties, which align with the assumptions made in the development of symmetry-informed reinforcement learning. Online training simulation shows that both DDPG-SDA accelerate the convergence toward a suboptimal policy compared to DDPG. In operation phase, the advantage of the symmetry-informed RL lies in improving tracking control performance in the regions of the state space that lack exploration. This comes from the improved coverage of the state space using symmetric samples. Although this advantage diminishes as exploration is enhanced, symmetric data augmentation (SDA) still highlights the potential to reduce exploration costs while and improving controller performance. This benefit stems from the ability of SDA to effectively capture symmetric motions in the aircraft model and provides informative samples.


\clearpage
\appendix

\section{Proof of Theorem 1}\label{Appendix_proof_of_theorem_1}

\setcounter{myTheo}{0}
\begin{myTheo}
    (Symmetry of $x_{k+1}$) Select two samples from the system \ref{nonlinear_system}, denoted as $(x_{k},a_{k},x_{k+1})$, $(x^{\prime}_{k},a^{\prime}_{k},x^{\prime}_{k+1})$, and a reference state $x=x^{*}$. Assuming Equations \ref{symxt}, \ref{symat} hold, then $x_{k+1},x^{\prime}_{k+1}$ are symmetric with respect to $x^{*}$ when\\
(1) $x^{*}=0\in\mathbb{R}^{n},G(x_{k})=G(x^{\prime}_{k}),F(x_{k})=F(x^{\prime}_{k})$\\
(2) $x^{*}\neq0\in\mathbb{R}^{n}, G(x_{k})=G(x^{\prime}_{k}), F(x_{k})=F(x^{\prime}_{k})=I\in\mathbb{R}^{n\times n}$.
\end{myTheo}

\begin{proof}
    Assume states $x_{k},x^{\prime}_{k}$ are symmetric to the reference $x=x^{*}$:
\begin{equation}\label{symmetryofx0}
    \frac{(x_{k}+x^{\prime}_{k})}{2} =x^{*}
\end{equation}

and $a_{k},a^{\prime}_{k}$ are symmetric to 0 $\in\mathbb{R}^{n}$:
\begin{equation}
    a_{k} = -a^{\prime}_{k}
\end{equation}

The predicted state using the model \ref{nonlinear_system}, when taking action $a_{k}$ at state $x_{k}$, is
\begin{equation}\label{xt+1}
        x_{k+1} = F(x_{k})x_{k} + G(x_{k}) a_{k}
\end{equation}

The predicted state using the model \ref{nonlinear_system}, when taking action $-a_{k}$ at state $x^{\prime}_{k}$, is
\begin{equation}\label{xt+1prime}
        x^{\prime}_{k+1} = F(x^{\prime}_{k})x^{\prime}_{k} - G(x^{\prime}_{k}) a_{k}
\end{equation}

From \ref{xt+1}, \ref{xt+1prime}, one has
\begin{equation}\label{symmetryofx1}
\begin{aligned}
        \frac{x_{k+1}+x^{\prime}_{k+1}}{2} =& \frac{1}{2}\left[F(x_{k} )x_{k}+G(x_{k}) a_{k}\right] + \frac{1}{2}\left[F(x^{\prime}_{k})x^{\prime}_{k}-G(x^{\prime}_{k}) a_{k}\right]\\
    =& \frac{(x_{k}+x^{\prime}_{k})}{2}+\frac{1}{2}\left[F(x_{k})x_{k}+G(x_{k})a_{k}-x_{k}\right]\\
    &+\frac{1}{2}\left[F(x^{\prime}_{k})x^{\prime}_{k}-G(x^{\prime}_{k}) a_{k}-x^{\prime}_{k}\right]
\end{aligned}
\end{equation}

Substitute \ref{symmetryofx0} into \ref{symmetryofx1}:
\begin{equation}\label{symmetryplane3}
\begin{aligned}
    \frac{x_{k+1}+x^{\prime}_{k+1}}{2} = x^{*}+&\frac{1}{2}[F(x_{k})x_{k}+G(x_{k}) a_{k}-x_{k}]\\
    +&\frac{1}{2}[F(x^{\prime}_{k})x^{\prime}_{k}-G(x^{\prime}_{k}) a_{k}-x^{\prime}_{k}]
\end{aligned}
\end{equation}

If $x_{k+1},x^{\prime}_{k+1}$ are symmetric with respect to $x^{*}$, the following equation should hold
\begin{equation}\label{symmetryplane4}
    \frac{x_{k+1}+x^{\prime}_{k+1}}{2}= x^{*}
\end{equation}

Comparing \ref{symmetryplane3} and \ref{symmetryplane4}, one has
\begin{equation}\label{generalcondition}
    \frac{1}{2}\left[F(x_{k})x_{k}+G(x_{k}) a_{k}-x_{k}\right]+\frac{1}{2}\left[F(x^{\prime}_{k})x^{\prime}_{k}-G(x^{\prime}_{k}) a_{k}-x^{\prime}_{k}\right]=0
\end{equation}

Rewrite \ref{generalcondition} as
\begin{equation}\label{generalcondition1}
\begin{aligned}
        \underbrace{\frac{1}{2}\left[F(x_{k})-I\right]x_{k}+\frac{1}{2}\left[F(x^{\prime}_{k})-I\right]x^{\prime}_{k}}_{\text{term1}}+\underbrace{\frac{1}{2}G(x_{k}) a_{k}-\frac{1}{2}G(x^{\prime}_{k}) a_{k} = 0}_{\text{term2}}
\end{aligned}
\end{equation}

Equation \ref{generalcondition1} provides the condition for $x_{k+1}$ and $x^{\prime}_{k+1}$ to be symmetric with respect to $x^{*}$. Terms 1 and 2 describe how the functions $F(x_{k})$ and $G(x_{k})$ influence the symmetry when $x_{k}$ and $x^{\prime}_{k}$ transition to $x_{k+1}$ and $x^{\prime}_{k+1}$. If both term 1 and term 2 are equal to 0, then the symmetry of $x_{k+1}$ and $x^{\prime}_{k+1}$ is the same as that of $x_{k}$ and $x^{\prime}_{k}$. If either term 1 or term 2 is nonzero, then $x_{k+1}$ and $x^{\prime}_{k+1}$ are no longer symmetric.

For further discussion, we consider two cases:

\textit{Case 1}: $x^{*}=0\in\mathbb{R}^{n}$, i.e. the reference state is zero. 

From \ref{symmetryofx0}, one has
\begin{equation}\label{x1kx2k}
    x_{k}=-x^{\prime}_{k}
\end{equation}

Substitute \ref{x1kx2k} into term 1:
\begin{equation}
\begin{aligned}
        &\frac{1}{2}\left[F(x_{k})-I\right]x_{k}+\frac{1}{2}\left[F(x^{\prime}_{k})-I\right]x^{\prime}_{k}\\
        =& \frac{1}{2}\left[F(x_{k})-I\right]x_{k}-\frac{1}{2}\left[F(x^{\prime}_{k})-I\right]x_{k}\\
        =&\frac{1}{2}\left[F(x_{k})-F(x^{\prime}_{k})\right]x_{k}
\end{aligned}
\end{equation}\\
Because $x_{k}=0\in\mathbb{R}^{n}$ does not always hold, the condition that term 1 to be identically 0 becomes $ F(x_{k})=F(x^{\prime}_{k})$.

Rewrite term 2 as
\begin{equation}
\begin{aligned}
        &\frac{1}{2}G(x_{k}) a_{k}-\frac{1}{2}G(x^{\prime}_{k}) a_{k} \\
       =&\frac{1}{2}\left[G(x_{k}) -G(x^{\prime}_{k})\right]a_{k}
\end{aligned}
\end{equation}

Because $a_{k}=0$ does not always hold, the condition that term 2 to be identically 0 becomes $ G(x_{k})=G(x^{\prime}_{k})$.



\textit{Case 2}: $x^{*}\neq 0\in\mathbb{R}^{n}$, i.e. the reference state is nonzero. 

From Equation \ref{symmetryofx0}, one has
\begin{equation}\label{x1x2isnot0}
    x_{k} = 2 x^{*} - x^{\prime}_{k}
\end{equation}

Substitute \ref{x1x2isnot0} to term 1:
\begin{equation}\label{xstarneq0}
\begin{aligned}
        &\frac{1}{2}\left[F(x_{k})-I\right]x_{k}+\frac{1}{2}\left[F(x^{\prime}_{k})-I\right]x^{\prime}_{k}\\
        =& \frac{1}{2}\left[F(x_{k})-I\right]x_{k}+\frac{1}{2}\left[F(x^{\prime}_{k})-I\right](2x^{*}-x_{k})\\
         =& \frac{1}{2}\left[F(x_{k})-I\right]x_{k}-\frac{1}{2}\left[F(x^{\prime}_{k})-I\right]x_{k}+\left[F(x^{\prime}_{k})-I\right]x^{*}\\
         =& \frac{1}{2}\left[F(x_{k})-F(x^{\prime}_{k})\right]x_{k}+\left[F(x^{\prime}_{k})-I\right]x^{*}\\
\end{aligned}
\end{equation}

Because $x^{*}\neq 0\in\mathbb{R}^{n}$, the condition for \ref{xstarneq0} equals 0 is

\begin{equation}
\begin{aligned}
    F(x_{k}) = I, F(x^{\prime}_{k}) = I
\end{aligned}
\end{equation}

The analysis of term 2 is similar to that in case 1.

Rewrite term 2 as
\begin{equation}
\begin{aligned}
    \frac{1}{2}G(x_{k}) a_{k}-\frac{1}{2}G(x^{\prime}_{k}) a_{k} =\frac{1}{2}\left[G(x_{k}) -G(x^{\prime}_{k})\right]a_{k}
\end{aligned}
\end{equation}

Because $a_{k}=0\in\mathbb{R}^{n}$ does not always hold, the condition that term 2 to be identically 0 becomes $ G(x_{k})=G(x^{\prime}_{k})$.
\end{proof}

\section{Implementation of two-step approximate value iteration}\label{appendix_API_gradient}
In two-step approximate value iteration, the critics are trained using minibatches sampled from the replay buffers $\mathcal{D}_{1}$ and $\mathcal{D}_{2}$, respectively. The actor is trained twice in each training round by each critic to ensure that both the explored and augmented samples are incorporated into the policy optimization. Target networks are introduced to stabilize the learning process through delayed parameter updates \cite{Zhang2021}.

The \textit{first-step} approximate value iteration is conducted on explored samples. For a minibatch $\mathcal{B}_{1}$ from the replay buffer $\mathcal{D}_{1}$, the Mean Square Error (MSE) loss is calculated as

\begin{equation}\label{losscritic1}
    L_{\text{critic}} = \frac{1}{N}\sum_{j=1}^{N}\delta_{j}^{2}
\end{equation}\\
where $N$ is the number of samples in $\mathcal{B}_{1}$, $j$ is the index of samples, $\delta_{j}$ is the Bellman residual on the $j$th sample:

\begin{equation}\label{bellmaninnovation}
    \delta_{j} = r^{j}_{k} + \gamma (\hat{Q}^{h})^{i}_{\psi_{1\text{target}}}(x^{j}_{k+1},(\mu_{\vartheta_{\text{target}}})^{i}(x^{j}_{k+1})) - (\hat{Q}^{h}_{\psi_{1}})^{i}(x^{j}_{k},a^{j}_{k})
\end{equation}\\
where $\psi_{1\text{target}}$ is the parameter set of the target critic.

The gradient of loss $L_{\text{critic}}$ with respect to parameter set $\psi_{1}$ is derived as

\begin{equation}
\begin{aligned}
        \frac{\partial L_{\text{critic}}}{\partial \psi^{i}_{1}} = &\frac{1}{N}\left(2\delta_{1}\frac{\partial \delta_{1}}{\partial \psi^{i}_{1}}+2\delta_{2}\frac{\partial \delta_{2}}{\partial \psi^{i}_{1}}+\dots+2\delta_{N}\frac{\partial \delta_{N}}{\partial \psi^{i}_{1}}\right)\\
        =& -\frac{1}{N}\Bigg(2\delta_{1}\frac{\partial (\hat{Q}^{h}_{\psi_{1}})^{i}(x^{1}_{k},a^{1}_{k})}{\partial \psi^{i}_{1}}+2\delta_{2}\frac{\partial (\hat{Q}^{h}_{\psi_{1}})^{i}(x^{2}_{k},a^{2}_{k})}{\partial \psi^{i}_{1}}+\\
        &\dots+2\delta_{N}\frac{\partial (\hat{Q}^{h}_{\psi_{1}})^{i}(x^{N}_{k},a^{N}_{k})}{\partial \psi^{i}_{1}}\Bigg)\\
        =&-\frac{2}{N}\sum^{N}_{j=1}\delta_{j}\frac{\partial (\hat{Q}^{h}_{\psi_{1}})^{i}(x^{j}_{k},a^{j}_{k})}{\partial \psi^{i}_{1}}
\end{aligned}
\end{equation}

The parameter set $\psi_{1}$ is updated as
\begin{equation}
    \psi^{i+1}_{1} = \psi^{i}_{1} + \eta_{\text{critic}} \frac{\partial L_{\text{critic}}}{\partial \psi^{i}_{1}}
\end{equation}

The parameter set $\psi_{1\text{target}}$ is updated as
\begin{equation}
    \psi^{i+1}_{1\text{target}} = \tau\psi^{i}_{1\text{target}}+(1-\tau)\psi^{i+1}_{1}
\end{equation}\\
where $0<\tau\leq 1$ is a delay factor.

The policy improvement is conducted according to the first critic. The MSE loss of the actor is 
\begin{equation}
    L_{\text{actor}} = \frac{1}{N}\sum^{N}_{j=1}l_{j}
\end{equation}\\
where $l_{j}$ is the estimated state-action value on $j$th sample.

\begin{equation}
    l_{j}=(\hat{Q}^{h}_{\psi_{1}})^{i+1}(x^{j}_{k},(\mu_{\vartheta})^{i}(x^{j}_{k}))
\end{equation}


The gradient of loss $L_{\text{actor}}$ with respect to actor parameter set $\vartheta^{i}$ is

\begin{equation}\label{lossgradientactor1}
    \frac{\partial L_{\text{actor}}}{\partial \vartheta^{i}}=\frac{1}{N}\left(\frac{\partial l_{1}}{\partial \vartheta^{i}}+\frac{\partial l_{2}}{\partial \vartheta^{i}}+\cdots+\frac{\partial l_{N}}{\partial \vartheta^{i}}\right)
\end{equation}\\
where
\begin{equation}
\begin{aligned}
        \frac{\partial l_{j}}{\partial \vartheta^{i}} &= \frac{(\partial \hat{Q}^{h}_{\psi_{1}})^{i+1}(x^{j}_{k},(\mu_{\vartheta})^{i}(x^{j}_{k}))}{\partial \vartheta_{i}}\\
    &=\frac{\partial (\hat{Q}^{h}_{\psi_{1}})^{i+1}(x^{j}_{k},(\mu_{\vartheta})^{i}(x^{j}_{k})}{\partial (\mu_{\vartheta})^{i}(x^{j}_{k})}\frac{\partial (\mu_{\vartheta})^{i}(x^{j}_{k})}{\partial \vartheta^{i}}
\end{aligned}
\end{equation}

The parameter set $\vartheta$ is updated once by
\begin{equation}
    \vartheta^{i+1} = \vartheta^{i} + \eta_{\text{actor}} \frac{\partial L_{\text{actor}}}{\partial \vartheta^{i}}
\end{equation}

The parameter set $\vartheta_{\text{target}}$ is updated once by
\begin{equation}
    \vartheta^{i+1}_{\text{target}} = \tau\vartheta^{i}_{\text{target}}+(1-\tau)\vartheta^{i+1}
\end{equation}

The \textit{second-step} approximate value iteration is conducted on augmented samples. For a minibatch $\mathcal{B}_{2}$ from buffer $\mathcal{D}_{2}$, the MSE loss of the second critic is 

\begin{equation}\label{losscritic2}
    L_{\text{critic}} = \frac{1}{N}\sum_{j=1}^{N}\delta_{j}^{2}
\end{equation}\\
where $N$ is the number of samples in  $\mathcal{B}_{2}$, $j$ is the index of samples, $\delta_{j}$ is the Bellman residual on the $j$th sample:

\begin{equation}\label{bellmaninnovation1}
        \delta_{j} = r^{j}_{k} + \gamma (\hat{Q}^{h})^{i}_{\psi_{2\text{target}}}(x^{j}_{k+1},(\mu_{\vartheta_{\text{target}}})^{i+1}(x^{j}_{k+1})) - (\hat{Q}^{h}_{\psi_{2}})^{i}(x^{j}_{k},a^{j}_{k})
\end{equation}\\
where $\psi_{2\text{target}}$ is the parameter set of the second target critic.

The gradient of loss $L_{\text{critic}}$ with respect to the parameter set $\psi_{2}$ is

\begin{equation}
\begin{aligned}
        \frac{\partial L_{\text{critic}}}{\partial \psi^{i}_{2}} =& \frac{1}{N}\left(2\delta_{1}\frac{\partial \delta_{1}}{\partial \psi^{i}_{2}}+2\delta_{2}\frac{\partial \delta_{2}}{\partial \psi^{i}_{2}}+\dots+2\delta_{N}\frac{\partial \delta_{N}}{\partial \psi^{i}_{2}}\right)\\
        =& -\frac{1}{N}\Bigg(2\delta_{1}\frac{\partial (\hat{Q}^{h}_{\psi_{2}})^{i}(x^{1}_{k},a^{1}_{k})}{\partial \psi^{i}_{2}}+2\delta_{2}\frac{\partial (\hat{Q}^{h}_{\psi_{2}})^{i}(x^{2}_{k},a^{2}_{k})}{\partial \psi^{i}_{2}}+\\
        &\dots+2\delta_{N}\frac{\partial (\hat{Q}^{h}_{\psi_{2}})^{i}(x^{N}_{k},a^{N}_{k})}{\partial \psi^{i}_{2}}\Bigg)\\
        =&-\frac{2}{N}\sum^{N}_{j=1}\delta_{j}\frac{\partial (\hat{Q}^{h}_{\psi_{2}})^{i}(x^{j}_{k},a^{j}_{k})}{\partial \psi^{i}_{2}}
\end{aligned}
\end{equation}

The actor is updated as
\begin{equation}
    \psi^{i+1}_{2} = \psi^{i}_{2} + \eta_{\text{critic}} \frac{\partial L}{\partial \psi^{i}_{2}}
\end{equation}

The target actor is updated as
\begin{equation}
    \psi^{i+1}_{2\text{target}} = \tau\psi^{i}_{2\text{target}}+(1-\tau)\psi^{i+1}_{2}
\end{equation}

The policy improvement is conducted by the second critic. Define the MSE loss as
\begin{equation}
    L_{\text{actor}} = \frac{1}{N}\sum^{N}_{j=1}l_{j}
\end{equation}\\
where $l_{j}$ is the estimated state-action value on $j$th sample.

\begin{equation}
    l_{j}=(\hat{Q}^{h}_{\psi_{2}})^{i+1}(x^{j}_{k},(\mu_{\vartheta})^{i+1}(x^{j}_{k}))
\end{equation}


The gradient of loss $L_{\text{actor}}$ with respect to parameter set $\vartheta$ is derived as

\begin{equation}\label{lossgradientactor2}
    \frac{\partial L_{\text{actor}}}{\partial \vartheta^{i+1}}=\frac{1}{N}\left(\frac{\partial l_{1}}{\partial \vartheta^{i+1}}+\frac{\partial l_{2}}{\partial \vartheta^{i+1}}+\cdots+\frac{\partial l_{N}}{\partial \vartheta^{i+1}}\right)
\end{equation}\\
where
\begin{equation}
\begin{aligned}
        \frac{\partial l_{j}}{\partial \vartheta^{i+1}} &= \frac{\partial (\hat{Q}^{h}_{\psi_{1}})^{i+1}(x^{j}_{k},(\mu_{\vartheta})^{i}(x^{j}_{k}))}{\partial \vartheta_{i+1}}\\
    &=\frac{\partial (\hat{Q}^{h}_{\psi_{1}})^{i+1}(x^{j}_{k},(\mu_{\vartheta})^{i+1}(x^{j}_{k}))}{\partial (\mu_{\vartheta})^{i+1}(x^{j}_{k})}\frac{\partial (\mu_{\vartheta})^{i+1}(x^{j}_{k})}{\partial \vartheta^{i+1}}
\end{aligned}
\end{equation}

The parameter set $\vartheta$ is updated for the second time as
\begin{equation}
    \vartheta^{i+2} = \vartheta^{i+1} + \eta_{\text{actor}} \frac{\partial L_{\text{actor}}}{\partial \vartheta^{i+1}}
\end{equation}

The parameter set $\vartheta_{\text{target}}$ is updated for the second time as
\begin{equation}
    \vartheta^{i+2}_{\text{target}} = \tau\vartheta^{i+1}_{\text{target}}+(1-\tau)\vartheta^{i+2}
\end{equation}

\bibliographystyle{elsarticle-num}
\bibliography{cas-refs}

\end{document}